%% file: iclr2026_conference.tex
\documentclass{article} 
\usepackage{iclr2026_conference,times}

\input{math_commands.tex}

\usepackage{algorithm}
\usepackage{algpseudocodex}
\usepackage{amsmath}
\usepackage{amssymb}
\usepackage{multirow}
\usepackage{booktabs}
\usepackage{makecell}
\usepackage{subcaption}
\usepackage{fancybox}
\usepackage{fancyvrb}
\usepackage{bm}
\usepackage{multirow}
\usepackage{booktabs}
\usepackage{bbding}
\usepackage{makecell}
\usepackage{subcaption}
\usepackage{fancybox}
\usepackage{fancyvrb}
\usepackage{colortbl}
\usepackage{xcolor}
\usepackage{graphicx} 
\usepackage{hyperref}
\usepackage{calligra}
\usepackage{url}
\usepackage{hyperref}
\usepackage{url}
\usepackage{wrapfig}

\title{TriVLA: A Triple-System-Based Unified \\ Vision-Language-Action Model with Episodic World Modeling for General Robot Control}

\author{Zhenyang Liu$^{1,2}$, Yongchong Gu$^{1,2}$, Sixiao Zheng$^{1,2}$, Yanwei Fu$^{1,2\dagger}$, Xiangyang Xue$^{1\dagger}$, Yu-Gang Jiang$^{1\dagger}$\vspace{0.3em}\\
$^{1}$ Fudan University \hspace{1.0em} $^{2}$ Shanghai Innovation Institute \vspace{0.3em}\\
\tt\small{lzyzjhz@163.com, yongchonggu22@m.fudan.edu.cn,}\\ 
\tt\small{
\{sxzheng18,yanweifu,xyxue,ygj\}@fudan.edu.cn}}

\iclrfinalcopy 
\begin{document}

\def\thefootnote{$^{\dagger}$}\footnotetext{Corresponding authors}
\def\thefootnote{\arabic{footnote}}
\maketitle
\begin{figure}[h]
	\centering
    \includegraphics[width=\linewidth]{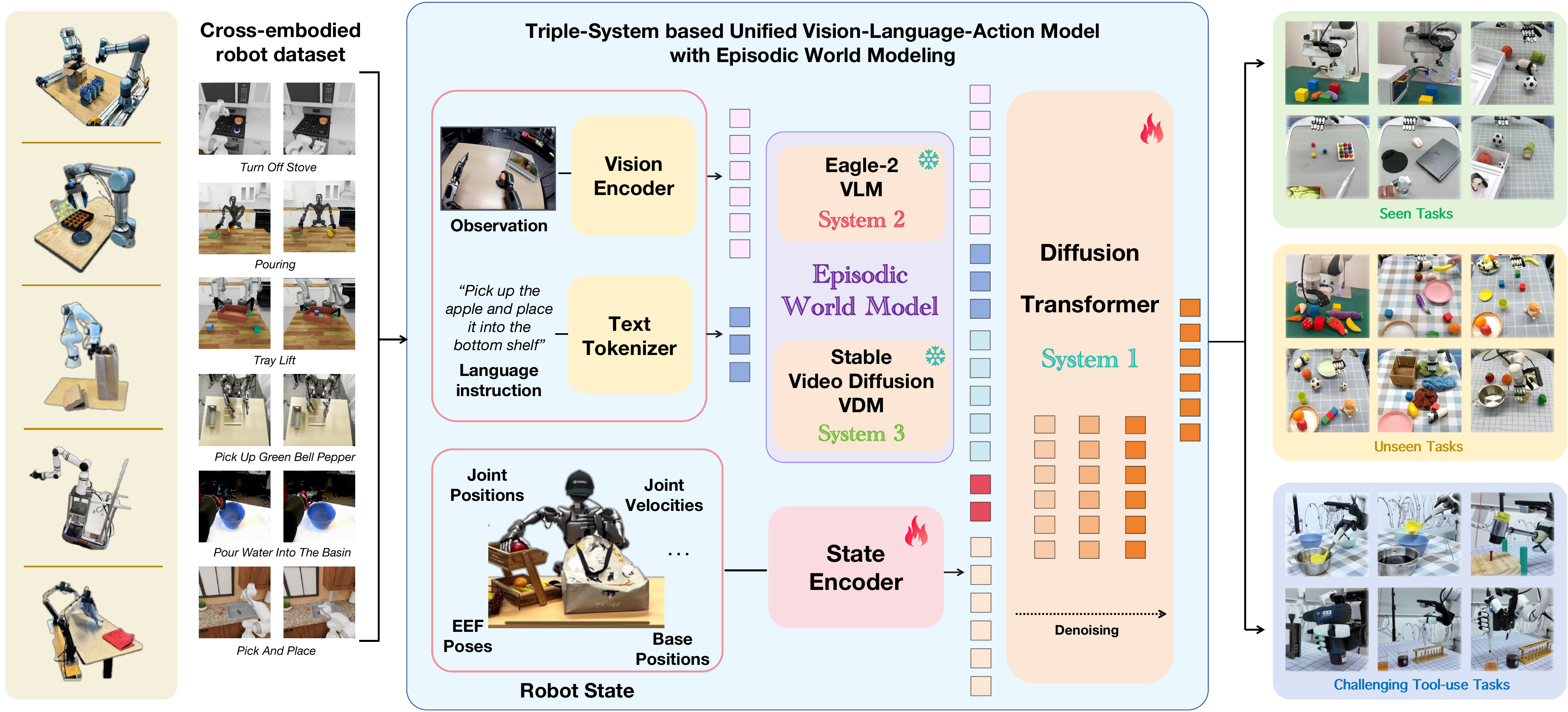}
	\caption{\textbf{TriVLA} is a unified Vision-Language-Action framework that adopts a triple-system architecture inspired by the \textbf{episodic world model}. Image and language inputs are processed by a Vision-Language Model for multimodal perception. A Video Diffusion Model provides dynamic world modeling and future prediction. The policy module integrates sequential outputs, robot state, and action history and generates real-time actions for complex manipulation tasks.
 \label{fig:teaser} }
\end{figure}

\begin{abstract}
Recent advances in vision–language models (VLMs) have enabled robots to follow open-ended instructions and demonstrate impressive commonsense reasoning. However, current vision–language–action (VLA) frameworks primarily rely on static representations and limited temporal context, restricting agents to short-horizon, reactive behaviors and hindering robust generalization in dynamic embodied environments.
Inspired by cognitive neuroscience theories of episodic memory, we are, to our knowledge, among the first to introduce a formalized episodic world model in VLA, enabling embodied robots to accumulate, recall, and predict sequential experiences.
As an instantiation of this concept, our unified \textbf{TriVLA} realizes the episodic world model through a triple-system architecture: integrating multimodal grounding from a pretrained VLM (System 2) and temporally rich dynamics perception from a video diffusion model (System 3). This enables the agent to accumulate and recall sequential experiences, interpret current contexts, and predict future environmental evolution.
Guided by episodic representations that span both the past and anticipated future, the downstream policy (System 1) generates coherent, context-aware action sequences through flow-matching and cross-modal attention mechanisms.
Experimental results show that TriVLA operates efficiently at ~36 Hz and consistently outperforms baseline models on standard benchmarks and challenging real-world manipulation tasks. It demonstrates strong long-horizon planning and open-ended intent understanding, showcasing the advantages of episodic world model-inspired reasoning for robust, generalizable robot intelligence. \textbf{Project Page:} \href{https://zhenyangliu.github.io/TriVLA/}{zhenyangliu.github.io/TriVLA}.
\end{abstract}

\section{Introduction}
\label{sec:intro}
\textit{“Episodic memory is the only memory system that allows mental time travel—backward into the past and forward into the future.”} 

\hfill --- Endel Tulving

Building on this cognitive foundation, we advocate that robotic agents, require an internal \textbf{episodic world model}: a representational system that not only recalls past interactions but also anticipates future dynamics, thereby enabling robust generalization in embodied environments.

Decades of cognitive neuroscience provide compelling evidence for this perspective. As first articulated by Tulving~\cite{tulving1972episodic}, episodic memory refers to the encoding, storage, and retrieval of experiences within their spatiotemporal context. This unique system empowers humans not only to recollect the past but also to simulate potential futures, thus grounding flexible planning and adaptive decision-making. Converging findings highlight the central roles of the hippocampus and prefrontal cortex in supporting episodic memory, enabling individuals to integrate sensory cues with temporal dynamics to construct predictive internal world models~\cite{tulving2002episodic,pritzel2017neural,blundell2016model,lin2018episodic,gershman2017reinforcement}. Episodic memory thus forms a fundamental component of intelligence, providing both the experiential basis for learning and the representational scaffolding for generalization across novel tasks.

Inspired by these insights, we propose the concept of an episodic world model: a unified framework that integrates multimodal grounding with temporally rich dynamic modeling. Unlike static scene representations, an episodic world model continuously accumulates, recalls, and predicts sequential experiences, equipping artificial agents with a memory system more akin to human intelligence. Recent advances in video diffusion models (VDMs)~\cite{blattmann2023stable, hong2022cogvideo, yang2024cogvideox, videoworldsimulators2024} provide a technological foundation for this paradigm, as they capture temporal continuity and physical dynamics across video sequences, naturally aligning with episodic memory principles and enabling richer, context-aware internal representations.

\begin{figure}[t]
	\centering
	\includegraphics[width=0.88\linewidth]{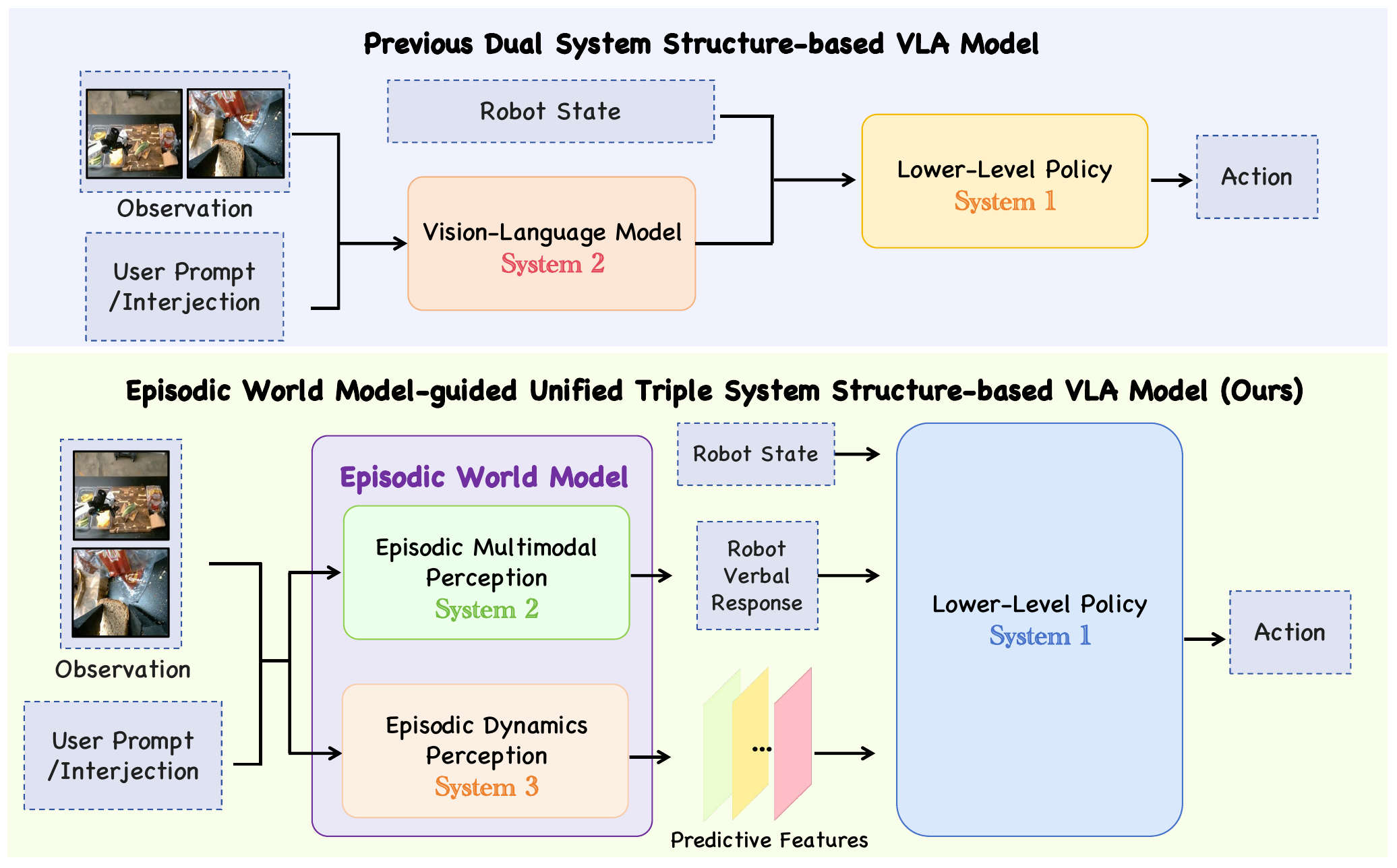}
	\caption{\textbf{Comparison between dual-system architectures and our episodic world model-guided TriVLA.} \textbf{TriVLA} implements the \textbf{episodic world model} using a triple-system architecture. In contrast, previous dual-system methods relied on static representations and limited temporal context, which restricted agents to short-horizon, reactive behaviors in dynamic environments.
 \label{sys}}
	\vspace{-0.2in}

\end{figure}

In parallel, vision–language models (VLMs)~\cite{liu2024visual, alayrac2022flamingo, li2023blip, zhang2023llama, bai2023qwen, gao2023llama, zhang2024mavis, zhang2024mathverse} have demonstrated impressive progress in instruction following and commonsense reasoning through large-scale pretraining on image–text corpora. Extending these capabilities, dual-system architectures have advanced VLMs into vision–language–action (VLA) models that generate action plans~\cite{ahn2022can, driess2023palm, huang2023voxposer, belkhale2024rt} and estimate SE(3) object poses~\cite{brohan2023rt, kim2024openvla, li2024manipllm}, enabling robots to map multimodal inputs into generalizable control behaviors. 
As illustrated in Figure~\ref{sys}, current VLM-based VLA systems~\cite{intelligence2025pi05visionlanguageactionmodelopenworld, black2024pi0visionlanguageactionflowmodel, brohan2023rt, kim2024openvla, pertsch2025fast} remain predominantly static: they depend on one or two instantaneous observations, overlooking the sequential and dynamic structures that characterize embodied interaction. As a result, they cannot encode or utilize temporally extended experiences, a capability similar to human episodic memory and crucial for robust performance in dynamic environments.

To bridge this gap, we introduce \textbf{TriVLA}, a unified Vision–Language–Action model that implements the episodic world model through a triple-system compositional architecture. Extending prior dual-system designs~\cite{bjorck2025gr00t, shi2025hi}, TriVLA explicitly integrates:
\begin{itemize}
    \item System 2: Episodic Multimodal Perception, a pretrained VLM that interprets observations and instructions, summarizing task goals and contextual cues.
    \item System 3: Episodic Dynamics Perception, a video diffusion model fine-tuned on large-scale human and robotic manipulation datasets~\cite{khazatsky2024droid, jin2024robotgpt, lu2024manigaussian}, which encodes sequences of past states and predicts future scene trajectories, realizing episodic context accumulation.
\end{itemize}
Together, Systems 2 and 3 jointly compose the episodic world model, fusing descriptive multimodal grounding with predictive temporal modeling. This joint representation empowers Policy Learning (System 1) to flexibly adapt actions based on accumulated experiences and anticipated future dynamics. 
By continually monitoring motion sequences embedded within the episodic world representation, the downstream policy induces an implicit inverse-dynamics prior~\cite{min2023trajectory, tian2024predictive}. 
This facilitates the transfer of the generalization capabilities inherent in the episodic world model to the robotic policy, meaning that the robot requires only a limited number of demonstrations to align its action space with the visual domain.
During training, System 1 utilizes action flow-matching and cross-attention mechanisms to integrate output tokens from both Systems 2 and 3. It adopts embodiment-specific encoders and decoders to manage variable state and action dimensions during motion generation. In addition, inspired by recent advances in robot learning, System 1 is designed to predict a chunk of actions rather than generating isolated actions at each timestep.

Our main contribution is an episodic world model inspired by cognitive neuroscience theories of episodic memory.
Building on this, we introduce a novel triple-system architecture within a unified Vision-Language-Action framework. The episodic world model that guides manipulation policy learning with temporally extended, context-aware signals.
As illustrated in Figure~\ref{fig:teaser}, this unified framework empowers robots to interpret complex prompts, reason over long horizons, and adapt adaptively in open-ended, dynamic scenarios.

Experimental results demonstrate that the proposed TriVLA consistently outperforms baseline algorithms, in both simulated~\cite{mees2022calvin,liu2024libero,yu2020meta} and real-world environments.
This highlights its effectiveness in aligning with human intent and achieving long-horizon task success.
Notably, TriVLA attains improvements of 0.21, 0.11, and 0.13 on the Calvin ABC→D, LIBERO, and MetaWorld benchmarks, respectively, compared to prior state-of-the-art methods.
In real-world experiments, TriVLA demonstrates strong effectiveness in dexterous hand manipulation tasks, particularly in long-horizon scenarios.

The contributions of this paper are summarized:
\begin{itemize}
    \item \textbf{\textit{Episodic World Model Inspired by Cognitive Neuroscience}}:    
    We propose an episodic world model for embodied agents, inspired by cognitive neuroscience theories of episodic memory. This model enables robots to accumulate, recall, and predict sequential multimodal experiences and offers a solid foundation for robust, adaptive control.
    \item \textbf{\textit{A Unified Triple-System Compositional Architecture}}: Building on this foundation, we present TriVLA, a triple-system architecture implementing the episodic world model. TriVLA provides high-level reasoning and dynamic prediction. Robots using TriVLA can understand complex prompts and perform long-horizon manipulation.
    \item \textbf{\textit{State-of-the-art Performance}}: TriVLA outperforms other baseline algorithms, including novel skill compositions beyond training combinations. This demonstrates the effectiveness in both alignment with human intent and long-horizon task success. 
\end{itemize}

\section{Related Work}

\paragraph{Vision-language-action models.}
Previous studies~\cite{ahn2022can, driess2023palm, huang2023voxposer, huang2024rekep} have advanced robotic comprehension of language and vision to autonomously generate task plans. Vision-language-action (VLA) models leverage VLMs' reasoning capabilities for SE(3) pose prediction: RT2~\cite{brohan2023rt} binarizes 7-DoF actions for autoregressive prediction; ManipLLM~\cite{li2024manipllm} incorporates affordance priors and chain-of-thought reasoning; OpenVLA~\cite{kim2024openvla} pretrains on Open X-Embodiment~\cite{o2023open} for improved generalization; and FAST~\cite{pertsch2025fast} introduces discrete cosine transform for efficient scalability.
Cognitively inspired dual-systems like GR00T N1~\cite{bjorck2025gr00t} and Hi Robot~\cite{shi2025hi} enhance adaptation to novel scenarios and accelerate task learning.
Additionally, various VLA approaches~\cite{liu2024robomamba, huang2024manipvqa, li2023vision, wu2023unleashing} achieve continuous action prediction by integrating policy heads (e.g., MLP, LSTM~\cite{graves2012long}) with regression losses in imitation learning.
Previous approaches are mostly static. They rely on current observations and ignore sequential dynamics. As a result, these methods cannot encode temporally extended experiences, which are crucial for robust performance in dynamic environments. In contrast, TriVLA uses a unified triple-system architecture. This design enables robots to interpret complex prompts and perform long-horizon manipulation tasks across various scenarios.

\paragraph{Future prediction in robotics.}
Prior studies have investigated leveraging future prediction to improve policy learning~\cite{bharadhwaj2024gen2act, chen2024igor, ye2024latent, guo2024prediction}.
SuSIE~\cite{black2023zero} bases its control policy on a predicted future keyframe produced by InstructPix2Pix~\cite{brooks2023instructpix2pix}, whereas UniPi~\cite{du2024learning} models inverse dynamics across two generated frames.
These approaches rely on a single-step future prediction to guide action selection, failing to fully capture the complexity of physical dynamics. Furthermore, denoising the final predicted image is time-consuming and results in reduced control frequency.
GR-1~\cite{wu2023unleashing} generates future frames and actions in an autoregressive manner. However, it produces only one image per forward pass, and its prediction quality trails behind diffusion-based techniques.
Seer~\cite{tian2024predictive} proposes an end-to-end framework that predicts actions by applying inverse dynamics models conditioned on forecasted visual states.
VPP~\cite{hu2024video} employs representations fine-tuned from video models to build a generalist robotic policy. In contrast, our TriVLA integrates Episodic Multimodal Perception and Episodic Dynamics Perception to provide both high-level reasoning and dynamic predictive representations. This enables prediction of sequential future frames while maintaining high-level reasoning, thereby more effectively guiding policy learning. 

\section{Preliminaries}
\noindent \textbf{Vision-language-action model.} 
Robotic manipulation is a longstanding core topic in robotics. Vision-language-action (VLA) models infer the robot’s next action—typically the end-effector pose—from visual input and human instructions. Recent advances in large pretrained vision-language models (VLMs) have enabled strong generalization across diverse, language-conditioned manipulation tasks. Most VLM-based VLA approaches adopt a dual-system architecture inspired by human cognition~\cite{kahneman2011thinking}, facilitating higher-level reasoning for complex, long-horizon task interpretation and action selection.
At each timestep $t$, the high-level system takes the captured image $\mathbf{o}_t$ from base and wrist-mounted cameras, along with the open-ended instruction $\mathbf{v}_t^{\text{in}}$, as input to generate reasoning tokens. 
The low-level policy uses these tokens, images, and robot states and outputs an action token sequence $\mathbf{v}_t^{\text{out}} \in \mathcal{V}^n$, where each action token represents a discrete bin of one dimension of the robot action space. The final robot action is extracted from this sequence using a post-processing function $f$, resulting in $\mathbf{a}_t = f(\mathbf{v}_t^{\text{out}})$. 

\section{Our Proposed TriVLA}\label{method}
\begin{figure*}
	\centering
    \includegraphics[width=0.90\linewidth]{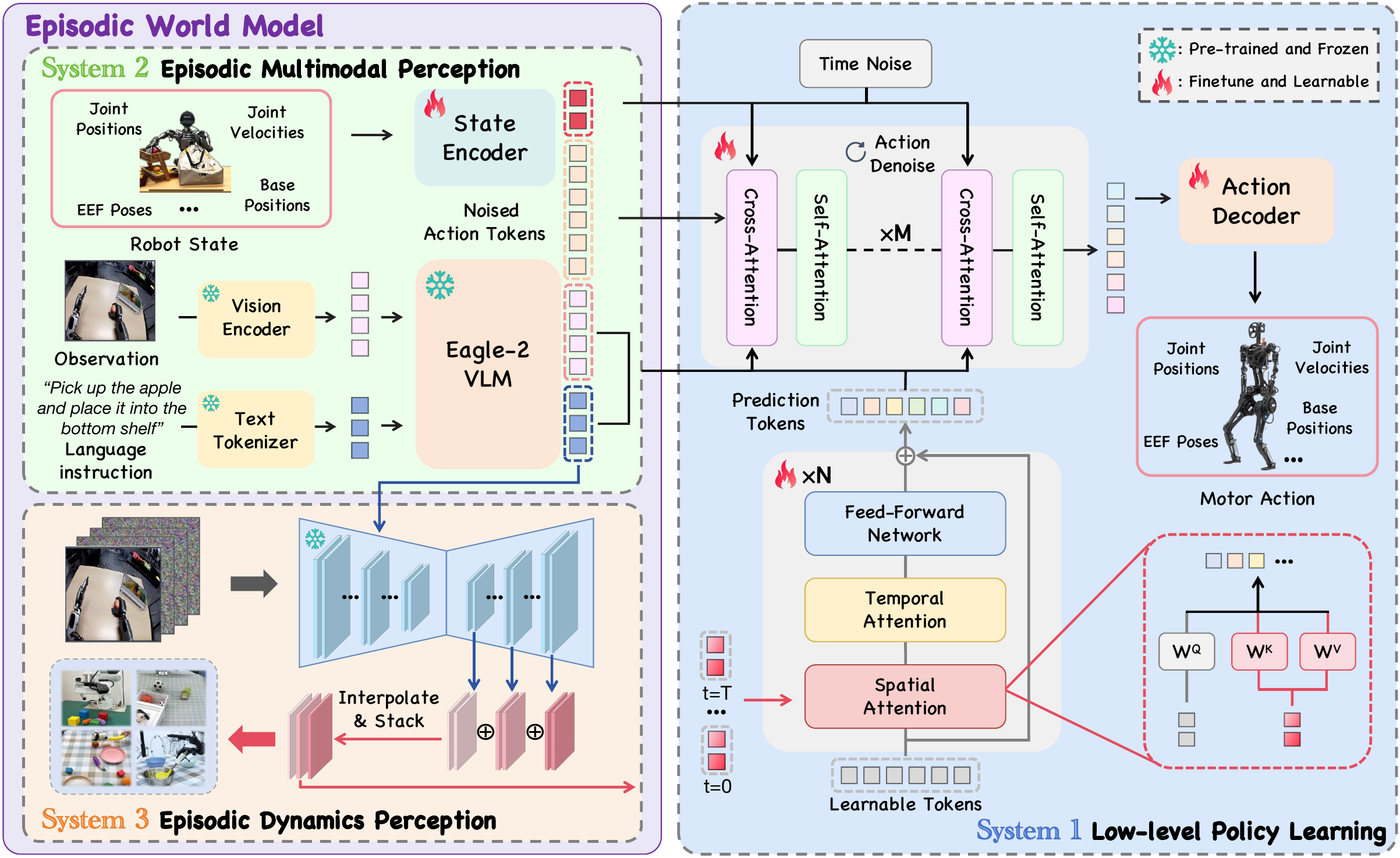}
	\caption{\textbf{The pipeline of TriVLA.} \textbf{TriVLA} is a unified Vision-Language-Action framework built on a triple-system paradigm. System 2 employs a pre-trained Eagle-2 VLM for episodic multimodal perception, while System 3 utilizes a general-purpose VDM to model episodic dynamics and sequential changes. Together, these modules form a joint episodic world model with rich, temporally extended representations. System 1 serves as the policy module, applying action flow-matching to integrate all outputs along with robot state and action history.
	\label{fig:overall} }	
	 \vspace{-0.2in}
\end{figure*}

TriVLA implements the episodic world model using a triple-system design (Figure~\ref{fig:overall}). It integrates (i) Episodic Multimodal Perception, which uses the Eagle-2 VLM~\cite{li2025eagle} to interpret visual inputs and language instructions, and (ii) Episodic Dynamics Perception, where a video diffusion model fine-tuned on large-scale manipulation datasets predicts future scene trajectories. These modules provide rich episodic context for policy generalization. The policy module leverages this context with flow matching~\cite{lipman2022flow} and a diffusion transformer (DiT) to generate actions.

\subsection{Episodic Multimodal Perception (System 2)}\label{ms1}
To realize this component, TriVLA employs the NVIDIA Eagle-2 vision–language model (VLM)~\cite{li2025eagle}, pretrained on large-scale internet corpora to jointly interpret observations and instructions, summarizing task goals and contextual cues. Eagle-2 is built from a SmolLM2 language model and a SigLIP-2 image encoder, aligned through a broad vision–language pretraining protocol. Images are encoded at a resolution of 224 × 224 pixels and then processed via pixel shuffle to produce 64 image token embeddings per frame. These image embeddings are jointly encoded with textual input by the LLM component of Eagle-2. The LLM and image encoder are aligned through a general vision-language pretraining protocol covering diverse tasks. During policy training, task descriptions and multiple images are input to the VLM following the chat format established during pretraining. The resulting vision-language tokens, denoted as \(Q_{vl}\) (batch size × sequence length × hidden dimension), are extracted from the LLM. Empirically, extracting embeddings from the 12th layer of the LLM, rather than from the final layer, yields faster inference and higher policy success rates. To handle varying robot state dimensions, TriVLA uses an embodiment-specific MLP to project each robot’s state into a shared embedding space, resulting in a state token \(Q_s \).

\subsection{Episodic Dynamics Perception (System 3)}\label{ms2}
To infuse extensive prior knowledge of dynamics into policy learning, we fine-tuned the 1.5B-parameter open-source Stable Video Diffusion (SVD) model~\cite{blattmann2023stable} as the Episodic Dynamics Perception module for robot manipulation. This video diffusion model, trained on large-scale human and robotic manipulation datasets, encodes sequences of past states and predicts future scene trajectories, enabling episodic context accumulation. By leveraging diverse sources—including internet human manipulation data, robot datasets, and self-collected data—the module robustly models sequential environmental changes essential for effective policy learning.
Then Episodic Dynamics Perception module $V_\theta$ is trained with diffusion objective, reconstructing the full video sequence $x_0 = s_{0:T}$ in dataset $D$ from noised samples $x_t=\sqrt{\bar\alpha_t}x_0+\sqrt{1 - \bar\alpha_t}\epsilon$:
\begin{equation}
\begin{split}
\label{eq:action_diff_loss}
\mathcal{L}_{D}
=\mathbb{E}_{x_0\sim D, \epsilon, t} \|V_\theta&(x_t,l_{emb},s_0)- x_0\|^2
\end{split}
\end{equation}
where $l_{emb}$ denotes the language feature from CLIP~\cite{radford2021learning}.Then we froze the fine-tuned Episodic Dynamics Perception module in downstream action learning.

However, denoising a complete video sequence is computationally intensive and may cause open-loop control problems, as highlighted in~\cite{du2024learning}.
Furthermore, videos in raw pixel format frequently contain abundant irrelevant information that can hinder effective decision-making.
To mitigate these challenges, we utilize the video diffusion model with a single forward pass.
Our key insight is that the initial forward step, despite not producing a clear video, offers a coarse trajectory of future states and informative guidance.
This observation is validated experimentally and illustrated in Figure~\ref{vp}.
Specifically, we concatenate the current image $s_0$ with the final noised latent $q(x_{t'} |x_0)$ (typically white noise) and input this combination into the System 2. The latent features are then directly utilized.
Previous work~\cite{xiang2023denoising} emphasizes that up-sampling layers in diffusion models produce more effective feature representations. The feature at the $m^{th}$ up-sampling layer, with width $W_m$ and height $H_m$, can be expressed as:
\begin{equation}
\begin{split}
\label{eq:2}
L_m = V_\theta(x_{t'},l_{emb},s_0)_{(m)},L_m\in\mathbb{R}^{T \times C_m \times W_m \times H_m}
\end{split}
\end{equation}
To efficiently integrate features from multiple up-sampling layers while eliminating manual layer selection, we propose an automatic feature aggregation approach across layers.
First, each layer’s feature map is linearly interpolated to a common height and width $W_p \times H_p$:
\begin{equation}
\begin{split}
\label{eq:3}
L_m'=\text{Interpolation}(L_m), L_m'\in \mathbb{R}^{T\times C_m\times W_p\times H_p}
\end{split}
\end{equation}
Subsequently, the features are concatenated along the channel dimension.
The final predictive visual representation $F_p \in \mathbb{R}^{T\times (\sum_m C_m)\times W_p\times H_p}$ is given by:
\[ F_p = \text{concate}((L_0',L_1',\dots,L_m'), dim=1)\]
For robots equipped with multiple camera perspectives, including third-person and wrist-mounted cameras, future states are predicted independently for each view, denoted as $F^{static}_p, F^{wrist}_p$. 

\subsection{Policy Learning Module (System 1)}\label{ms3}
Systems 2 and 3 together form the episodic world model, combining descriptive multimodal grounding with predictive temporal modeling.
This combined representation allows System 1 (Policy Learning) to flexibly adapt its actions using both accumulated experience and anticipated future dynamics.
The predictive representations generated by the video diffusion model remain high-dimensional because they encode sequences of image features.
To efficiently aggregate information across spatial, temporal, and multi-view dimensions, TriVLA compresses these representations into a fixed set of tokens.
We initializes learnable tokens $Q_{[0:T,0:L]}$ with fixed length $T \times L$, performing spatial-temporal attention \cite{blattmann2023align} on each corresponding frame, followed by feed-forward layers. 
Formally, this branch can be expressed as follows where $i$ is the index of frame:
\begin{equation}\label{eq:Spatial-attn}
\begin{split}
Q' = \{\text{Spat-Attn} (Q[i]&,(F^{static}_p[i],F^{wrist}_p[i]) ) \}_{i=0}^{T}\\ 
Q_{p} = \text{FFN}&(\text{Temp-Attn}(Q'))
\end{split}
\end{equation}
After the Episodic Multimodal Perception module (System 2) extracts vision-language tokens $Q_{vl}$, and the  Episodic Dynamics Perception module (System 3) aggregates future dynamic features into predictive tokens $Q_p$, a diffusion policy is employed as the action head to generate the action sequence $a_0 \in A$ conditioned on $Q_{vl}$ and $Q_p$.	
The aggregated tokens $Q_{vl}$ and $Q_p$ are integrated into the diffusion transformer blocks via cross-attention layers.	
The diffusion policy aims to reconstruct the original action $a_0$ from the noised action $a_k = \sqrt{\bar\beta_k}a_0 + \sqrt{1 - \bar\beta_k}\epsilon$, where $\epsilon$ denotes white noise and $\bar\beta_k$ is the noise coefficient at step $k$.	
This process can be interpreted as learning a denoiser $D_\psi$ to approximate the noise $\epsilon$.	
After the final DiT block, we apply an embodiment-specific action decoder $A_d$, implemented as a multi-layer perceptron (MLP). This decoder processes the final tokens to predict actions and minimize the following loss function:	
\begin{equation}
\begin{split}
\label{eq:action_diff_loss}
\mathcal{L}_{\text{diff}}(\psi; A)
=\mathbb{E}_{a_0, \epsilon, k}
   \|A_d(D_\psi(a_k,Q_{vl},Q_p))- a_0\|^2
\end{split}
\end{equation}


\section{Experiments}
\noindent\textbf{Simulation Benchmarks.} 
We evaluate {our method} on three widely used simulation benchmarks for long-horizon manipulation. CALVIN~\cite{mees2022calvin} assesses policies in the challenging ABC$\rightarrow$D generalization setting; following GR-1~\cite{wu2023unleashing}, we use only language-annotated ABC data for training and testing in the unseen D environment. LIBERO~\cite{liu2024libero} consists of four suites (Spatial, Object, Goal, Long), each with 10 tasks and 50 demonstrations. MetaWorld~\cite{yu2020meta,radosavovic2023real} includes 50 Sawyer-robot tasks.

\noindent \textbf{Real-world Experimental Setups.} We use a KINOVA GEN2 robot with a RealSense D455 depth camera mounted in an eye-to-hand configuration. In an indoor environment, we arranged various objects to encourage generalization of manipulation skills across different scenes. To further evaluate the episodic world model, we designed long-horizon, high-dynamic tasks that require the agent to accumulate, recall, and predict sequential multi-modal experiences. The RealSense D455 captured the entire scene and the robot’s state from both third-person and wrist perspectives.

\noindent \textbf{Comparison Methods.} Generalist robot policies have been extensively studied. We compare TriVLA with a representative subset of state-of-the-art and related methods, including RT-1~\cite{brohan2022rt} (semantic features with EfficientNet and FiLM), Diffusion Policy~\cite{chi2023diffusion} (conditional diffusion visuomotor policy), Robo-Flamingo~\cite{li2023vision} (LLM-visual integration via Flamingo~\cite{alayrac2022flamingo}), UniPi~\cite{du2024learning} (video prediction plus IK for action inference), MDT~\cite{reuss2024multimodaldiffusiontransformerlearning} (diffusion transformer), Susie~\cite{black2023zero} (goal image via InstructPix2Pix~\cite{brooks2023instructpix2pix} and diffusion learning), GR-1~\cite{wu2023unleashing} (autoregressive video and action policy), Robo-Uniview~\cite{liu2024robouniview} (3D-aware encoder), Vidman~\cite{wen2024vidman} (pre-trained video representation), Seer~\cite{tian2024predictive} (end-to-end vision-action via inverse dynamics), and VPP~\cite{hu2024video} (video diffusion temporal reasoning). These baselines encompass direct action learning, predictive modeling, and vision-language integration paradigms.

\begin{table*}[t]
\centering
\small
\caption{\textbf{Zero-shot long-horizon evaluation on the Calvin ABC$\rightarrow$D benchmark (Avg. Len).}}
\vspace{-2mm}
\label{table1}
\resizebox{\linewidth}{!}{
\begin{tabular}{cccccccccc}
\toprule
\multirow{2}{*}{\textbf{Category}}& \multirow{2}{*}{\textbf{Method}}& \multirow{2}{*}{\textbf{Annotated Data}} & \multicolumn{6}{c}{\textbf{$i^{th}$ Task Success Rate}} \\ \cline{4-9} 
 & & & \textbf{1} & \textbf{2} & \textbf{3} & \textbf{4} & \textbf{5} & \textbf{Avg. Len $\uparrow$} \\ \hline
\multirow{3}{*}{\begin{tabular}[c]{@{}c@{}}Direct Action \\ Learning Method\end{tabular}} 
& RT-1~ &100\%ABC & 0.533 & 0.222 & 0.094 & 0.038 & 0.013 & 0.90 \\
& Diffusion Policy &100\%ABC & 0.402 & 0.123 & 0.026 & 0.008 & 0.00 & 0.56 \\ 
& Robo-Flamingo &100\%ABC & 0.824 & 0.619 & 0.466 & 0.331 & 0.235 & 2.47 \\ \hline
3D Method & RoboUniview&100\%ABC & 0.942 & 0.842 & 0.734 & 0.622 & 0.507 & 3.65 \\ \hline
\multirow{8}{*}{\begin{tabular}[c]{@{}c@{}}Future Prediction \\ Related Method\end{tabular}} 
 & Uni-Pi &100\%ABC & 0.560 & 0.160 & 0.080 & 0.080 & 0.040 & 0.92 \\
& MDT&100\%ABC & 0.631 & 0.429 & 0.247 & 0.151 & 0.091 & 1.55 \\
 & Susie&100\%ABC & 0.870 & 0.690 & 0.490 & 0.380 & 0.260 & 2.69 \\
 & GR-1&100\%ABC & 0.854 & 0.712 & 0.596 & 0.497 & 0.401 & 3.06 \\ 
& Vidman&100\%ABC & 0.915 & 0.764 & 0.682 & 0.592 & 0.467 & 3.42 \\ 
& Seer &100\%ABC & 0.963 & 0.916 & 0.861 & 0.803 & 0.740 & 4.28 \\ 
& VPP &100\%ABC & 0.965 & 0.909 & 0.866 & 0.820 & 0.769 & 4.33 \\ 
& \cellcolor[HTML]{efefef}\textbf{TriVLA (ours)} &\cellcolor[HTML]{efefef}100\%ABC & \cellcolor[HTML]{efefef}\textbf{0.968} & \cellcolor[HTML]{efefef}\textbf{0.924} &\cellcolor[HTML]{efefef} \textbf{0.868} & \cellcolor[HTML]{efefef}\textbf{0.832} & \cellcolor[HTML]{efefef}\textbf{0.818} & \cellcolor[HTML]{efefef}\textbf{4.37} \\ \hline \midrule
\multirow{3}{*}{\begin{tabular}[c]{@{}c@{}}Data \\ Efficiency\end{tabular}}
& GR-1&10\%ABC & 0.672 & 0.371 & 0.198 & 0.108 & 0.069 & 1.41 \\ 
& VPP&10\%ABC & 0.878 & 0.746 & 0.632 & 0.540 & 0.453 & 3.25 \\ 
& \cellcolor[HTML]{efefef}\textbf{TriVLA (ours)} &\cellcolor[HTML]{efefef}10\%ABC &\cellcolor[HTML]{efefef}\textbf{0.914} & \cellcolor[HTML]{efefef}\textbf{0.768} & \cellcolor[HTML]{efefef}\textbf{0.644} & \cellcolor[HTML]{efefef}\textbf{0.564} & \cellcolor[HTML]{efefef}\textbf{0.512} & \cellcolor[HTML]{efefef}\textbf{3.46} \\ 
\bottomrule
\end{tabular}
}
\vspace{-2mm}

\end{table*}

\begin{table*}[t]
\centering
\small
\caption{\textbf{LIBERO benchmark experimental results.}}
\vspace{-6pt}
\begin{tabular}{lccccc}
\toprule
 & Average ($\uparrow$) & Spatial ($\uparrow$) & Object ($\uparrow$) & Goal ($\uparrow$) & Long ($\uparrow$) \\
\midrule
Diffusion Policy & 72.4 $\pm$ 0.7\% & 78.3 $\pm$ 1.1\% & 92.5 $\pm$ 0.7\% & 68.3 $\pm$ 1.2\% & 50.5 $\pm$ 1.3\% \\
Octo & 75.1 $\pm$ 0.6\% & 78.9 $\pm$ 1.0\% & 85.7 $\pm$ 0.9\% & 84.6 $\pm$ 0.9\% & 51.1 $\pm$ 1.3\% \\
OpenVLA & 76.5 $\pm$ 0.6\% & 84.7 $\pm$ 0.9\% & 88.4 $\pm$ 0.8\% & 79.2 $\pm$ 1.0\% & 53.7 $\pm$ 1.3\% \\
\rowcolor[HTML]{EFEFEF} TriVLA (ours) & \textbf{87.0 $\pm$ 0.7 \%}& \textbf{91.2 $\pm$ 0.8\%} & \textbf{93.8 $\pm$ 0.7\%} & \textbf{89.8 $\pm$ 0.9\%} & \textbf{73.2 $\pm$ 0.5\%}\\
\bottomrule
\end{tabular}
\vspace{-8pt}
\label{tab:libero_performance}
\end{table*}

\begin{figure}[t] \centering \includegraphics[width=0.85\linewidth]{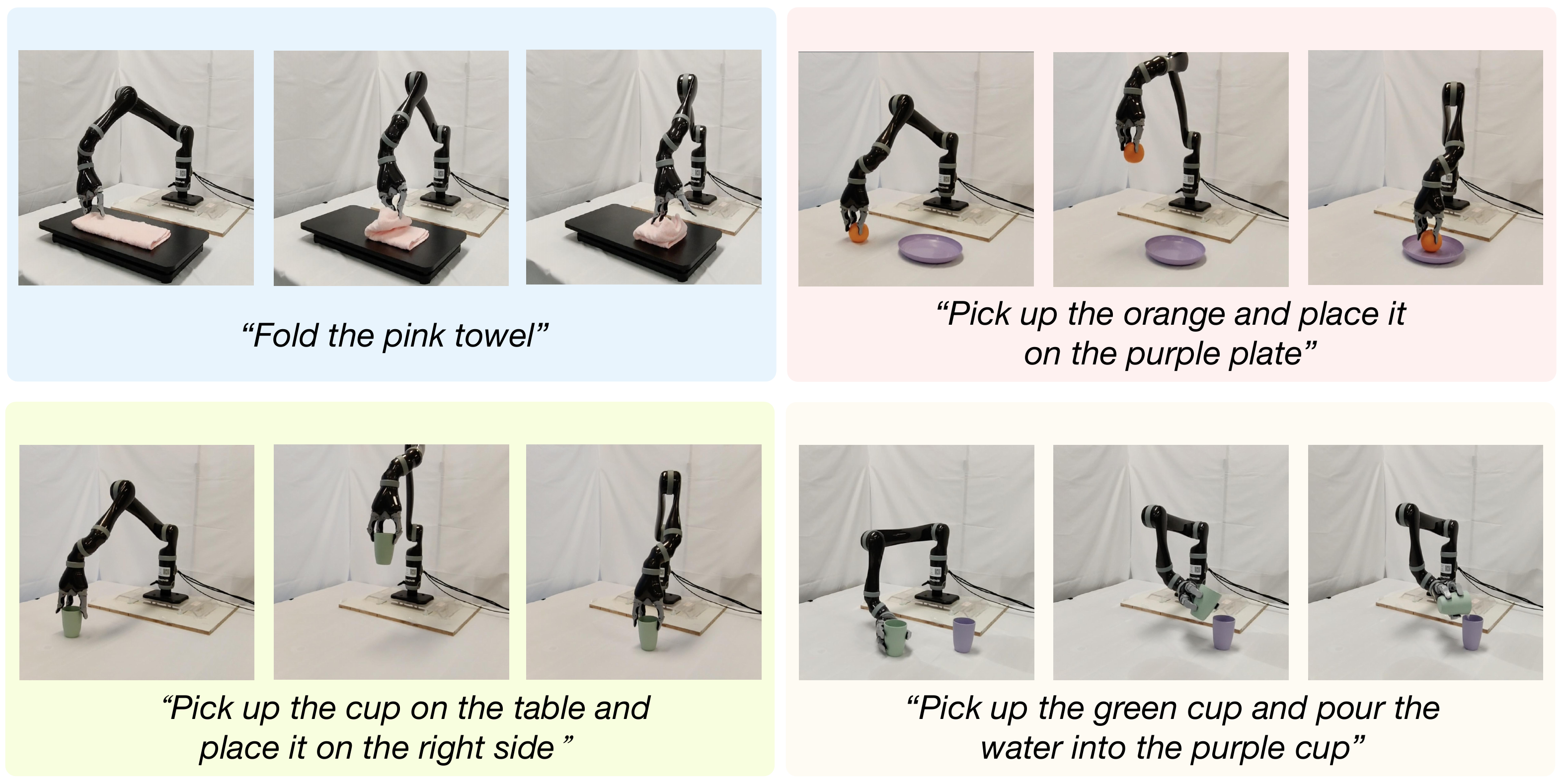} \vspace{-0.1in} \caption{\textbf{Qualitative case study of short-horizon tasks.} Our \textbf{TriVLA} performs well on short-horizon tasks. In the real-world tasks, it leverages a triple-system architecture that unifies Episodic Multimodal Perception and Dynamics Perception—both crucial for generalizable policy learning. \label{qual_sing}} \vspace{-1mm} \end{figure}

\begin{figure*}[t]
	\centering
	\includegraphics[width=\linewidth]{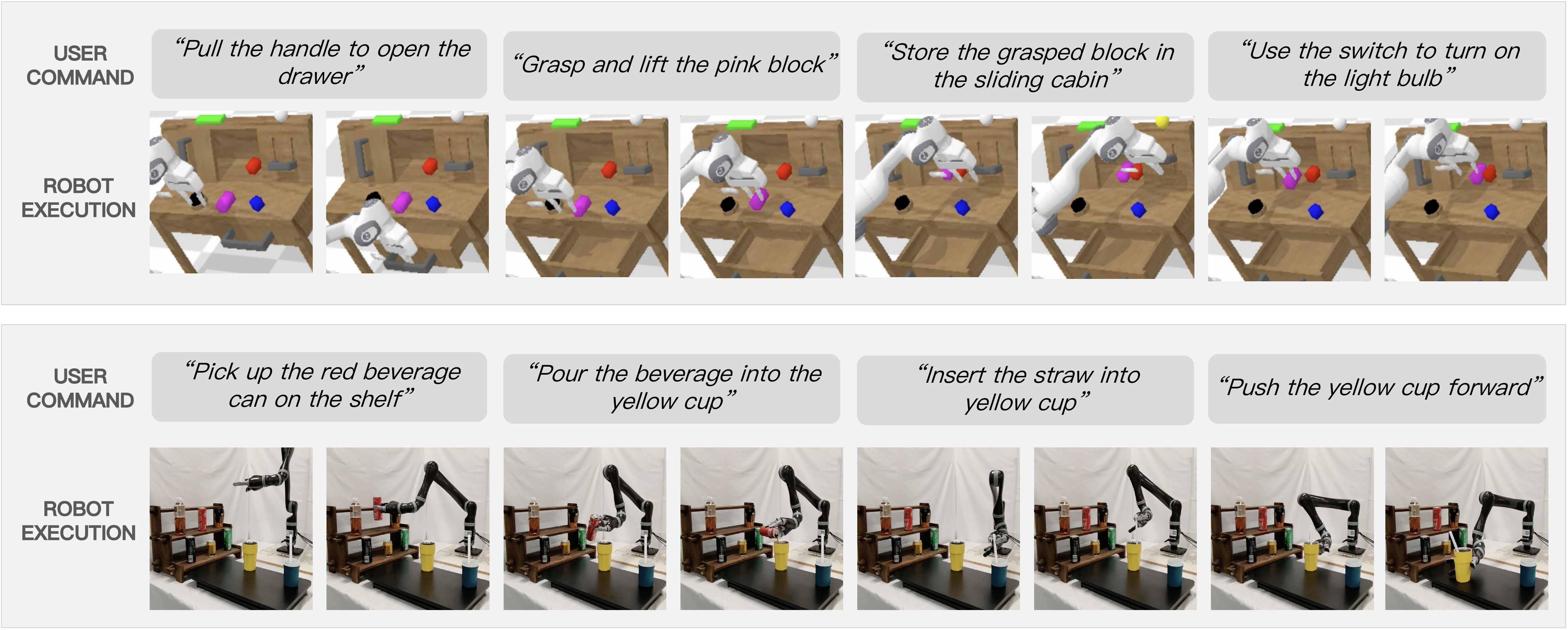}
    \vspace{-0.2in}
	\caption{\textbf{Qualitative results of long-horizon tasks.} Our \textbf{TriVLA} performs well on long-horizon tasks. In the CALVIN and real-world tasks, it leverages a triple-system architecture that unifies multiple systems for generalizable policy learning.
	\label{qual_1}}
    \vspace{-4mm}
\end{figure*}

\subsection{Experimental Results}
\noindent \textbf{Quantitative Results.} 
Comparisons on the Calvin benchmark are presented in Table \ref{table1}. 
Experimental results for Robo-Flamingo, Susie, GR-1, and 3D Diffuser Actors are extracted from their respective original publications.
MDT results are obtained from the official implementation. RT-1 and UniPi results are sourced from~\cite{li2023vision} and~\cite{black2023zero}, respectively.
We additionally executed the Diffusion Policy using the official open-source codebase with CLIP language conditioning. Our proposed TriVLA significantly outperforms previous state-of-the-art results.
Remarkably, despite training on only 10\% of the annotated Calvin ABC dataset, our method achieved an average task completion length of 3.46, outperforming related methods trained on the full dataset.
Furthermore, TriVLA attained the highest performance on the MetaWorld benchmark comprising 60 tasks, as detailed in Table~\ref{tab:meta}. 
TriVLA outperforms the strongest VPP baseline in average success rate.
\begin{wraptable}{r}{6.2cm}
\setlength{\tabcolsep}{5pt}
\small
\centering
\caption{\textbf{Performance on the MetaWorld.}}
\vspace{-6pt}
\scalebox{0.9}{
\begin{tabular}{lcccc}
\specialrule{1.5pt}{0pt}{0pt}
\textbf{Method} & \textbf{Easy} & \textbf{Middle} & \textbf{Hard} & \textbf{Avg~$\uparrow$} \\ \hline\hline
RT-1 & 0.603 & 0.030 & 0.014 & 0.331 \\
Diffusion Policy & 0.433 & 0.072 & 0.089 & 0.299 \\
Susie & 0.542 & 0.213 & 0.244 & 0.420 \\
GR-1 & 0.695 & 0.337 & 0.448 & 0.582 \\
VPP & 0.822 & 0.507 & 0.519 & 0.679 \\
\rowcolor[HTML]{EFEFEF}\textbf{TriVLA (ours)} & \textbf{0.857} & \textbf{0.528} & \textbf{0.563} & \textbf{0.714} \\
\specialrule{1.5pt}{0pt}{0pt}
\end{tabular}
\vspace{-12pt}
}
\label{tab:meta}
\end{wraptable}
Quantitative results on the LIBERO benchmark are presented in Table~\ref{tab:libero_performance},where each method is evaluated over 500 trials per task suite using 3 random seeds.
The results demonstrate that TriVLA effectively adapts to LIBERO simulation tasks, achieving the best or competitive performance.

\noindent \textbf{Qualitative Results.} 
Figure~\ref{qual_sing} and ~\ref{qual_1} shows two qualitative examples of action sequences in both simulation and the real world. Given multiple consecutive instructions, TriVLA can comprehend intent and leverage prediction to complete long-horizon tasks. These results show that TriVLA supports generalizable policy learning by integrating Episodic Multimodal Perception and Episodic Dynamics Perception. Robots using TriVLA can understand complex sequential prompts, reason across extended event horizons, and adapt to dynamic scenarios for effective long-horizon task execution.

\subsection{Ablation Study}
\begin{figure*}[t]
	\centering
	\includegraphics[width=\linewidth]{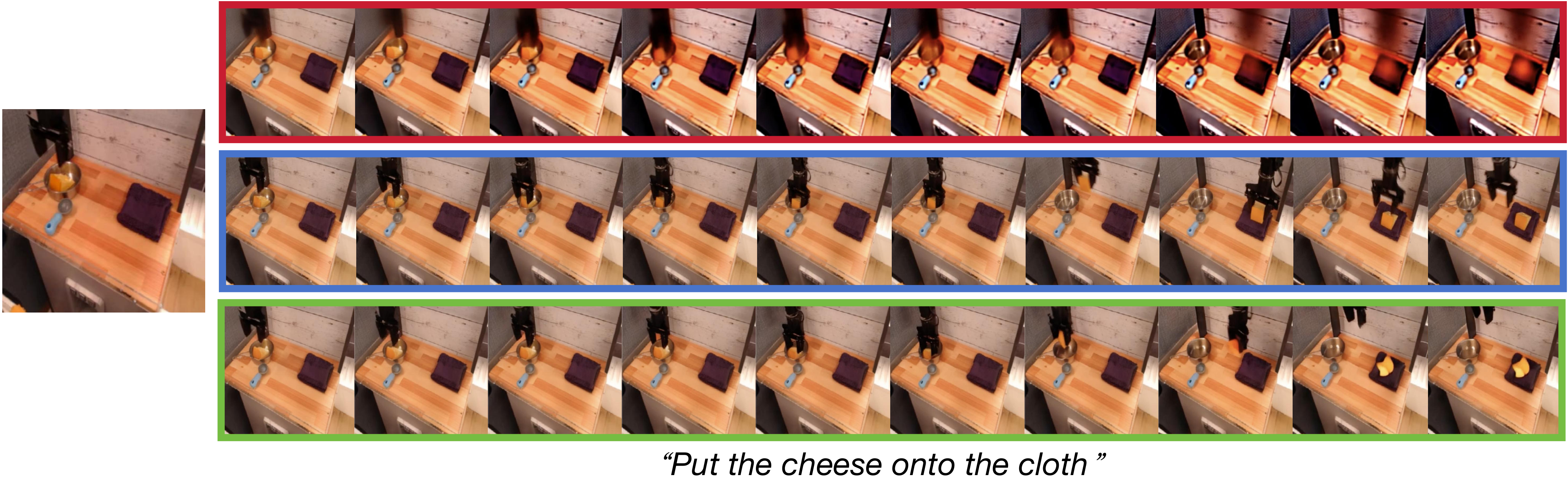}
    \vspace{-0.2in}
	\caption{{\textbf{Visualization of Episodic Dynamics Perception.} 
    The red box indicates one-step prediction, the blue box corresponds to full-step prediction, and the green box marks the ground truth.	
    \label{vp}}}
    \vspace{-0.15in}
\end{figure*}

\noindent \textbf{Visualization of Episodic Dynamics Perception.} We use a stable video diffusion model as the Episodic Dynamics Perception module. Its forward pass generates visual representations that capture both current scene information and predicted future dynamics over extended time horizons. Figure~\ref{vp} visualizes ground-truth futures, single-step predictions, and full-sequence predictions on the Bridge benchmark~\cite{walke2023bridgedata}. Results show that full-step predictions are reasonable, while single-step representations effectively capture key motion cues—such as object and robot arm movement—benefiting downstream action learning. Overall, the Episodic Dynamics Perception module can model entire video sequences and predict future frames conditioned on current observations and instructions, 
demonstrating strong understanding of physical dynamics.

\noindent \textbf{Effectiveness of Episodic Multimodal Perception Module.}
The System 2 Episodic Multimodal Perception module (EMP) of TriVLA is a pretrained Vision-Language Model (VLM) responsible for processing the robot’s visual perception and language instructions to interpret the environment and comprehend task goals.
Experiments were conducted to evaluate the effectiveness of System 2.
As summarized in Table~\ref{table:ablation}, integration of the Episodic Multimodal Perception module 
improved performance from 4.06 to 4.37, with inference time increasing from 136 ms to 155 ms.
\begin{wraptable}{r}{8.3cm}
\setlength{\tabcolsep}{4pt}
\small
\centering
\caption{\textbf{Sub-system ablation studies on the CALVIN.}}
\vspace{-6pt}
\scalebox{0.95}{
\begin{tabular}{ccc|ccc}
\specialrule{1.5pt}{0pt}{0pt}
EMP & EDP & L-Policy & Avg. Length~$\uparrow$ & Latency~$\downarrow$ & Params~$\downarrow$ \\
\hline\hline
 &  & \checkmark & 3.68 & 29.29ms  & 0.53B \\
 & \checkmark & \checkmark & 4.06 & 115.19ms & 1.87B \\
\checkmark & \checkmark & \checkmark & \textbf{4.37} & \textbf{142.69ms} & \textbf{3.39B} \\
\specialrule{1.5pt}{0pt}{0pt}
\end{tabular}
}
\vspace{-12pt}
\label{table:ablation}
\end{wraptable}
These results suggest that integrating System 2 significantly enhances the accuracy of action generation.

\noindent \textbf{Effectiveness of Episodic Dynamics Perception Module.} The System 3 Episodic Dynamics Perception module (EDP) is a general-purpose video diffusion model fine-tuned on internet-sourced human and robot manipulation datasets.
This module is designed to develop a controllable video generation model enhancing predictive capabilities within the manipulation domain.
To assess its effectiveness, we perform an ablation study by integrating the Episodic Dynamics Perception module into the proposed TriVLA framework.
The results, summarized in Table~\ref{table:ablation}, were obtained using a single NVIDIA H100 GPU.
Notably, incorporation of the Episodic Dynamics Perception results in a significant enhancement of performance.


\section{Conclusion}
TriVLA is the first framework to formalize an episodic world model within a unified triple-system architecture, drawing inspiration from cognitive neuroscience theories of episodic memory. By integrating multimodal grounding and rich temporal dynamics, TriVLA provides high-level reasoning and dynamic prediction, enabling robots to accumulate, recall, and predict sequential experiences.
Experiments show that TriVLA operates efficiently and consistently outperforms state-of-the-art policy baselines. TriVLA significantly improves long-horizon reasoning, sample efficiency, and open-ended goal achievement.
These results highlight the potential of episodic world model reasoning as a solid foundation for robust and generalizable robot control systems.

\section*{Ethics statement}
TriVLA advances vision–language–action (VLA) research by introducing an episodic world model that enables embodied agents to accumulate, recall, and predict sequential experiences for robust, long-horizon decision-making. This capability lowers barriers to building more generalizable and intelligent robots, which can benefit applications in assistive robotics, manufacturing, and human–robot interaction. At the same time, these advances raise ethical considerations regarding potential misuse (e.g., autonomous systems acting beyond intended safety boundaries, reinforcement of social or cultural biases in pretrained vision–language models) and broader societal impacts (e.g., displacement of human labor or over-reliance on autonomous decision-making). We encourage further research into transparent evaluation, safety alignment, and bias mitigation, as well as careful consideration of the ethical implications when deploying such systems in real-world settings. 

\section*{Reproducibility Statement}
We are committed to ensuring the reproducibility of all results reported in this work. Upon publication, we will release the TriVLA codebase, pretrained checkpoints for all system components (vision–language model, video diffusion model, and policy network), as well as training and evaluation scripts. Detailed descriptions of the episodic world model architecture, triple-system integration, and flow-matching mechanisms are provided in Supplemental Material~\ref{repro}, along with the hyperparameters and dataset preprocessing steps. We also specify all benchmarks, metrics, and experimental protocols used in both simulated and real-world evaluations. To facilitate independent verification, we will include environment setup instructions, hardware requirements, and example configurations, ensuring the seamless replication of our results.

\bibliography{iclr2026_conference}
\bibliographystyle{iclr2026_conference}

\newpage

\appendix
\section*{\textbf{Supplemental Material}}

To better understanding of this work, we offer additional details, analysis, and results as follows:

\begin{itemize}
   \item \textbf{A \;\; Use of Large Language Models (LLMs)} \\
    In this section, we clarify the use of LLMs in our paper, which primarily served as an assistants for refining and polishing the paper during the writing process.

    \item \textbf{B \;\; Implementation Details} \\
    In this section, we present the implementation details of TriVLA and its inherent Episodic World Model, including the training procedure and rollout process.

    \item \textbf{C \;\; Demo Video} \\
    In this section, we present the performance of TriVLA on short-horizon and long-horizon tasks to verify the practical effect of the Episodic World Model.

    \item \textbf{D \;\; Comparison Methods} \\
    In this section, we select a representative subset of prior methods for comparison. 
   
   \item \textbf{E \;\; Details and More Results of Episodic Dynamics Perception} \\
    In this section, we present detailed visualizations and results for Episodic Dynamics Perception in TriVLA. We employ a stable video diffusion model as the core module for Episodic Dynamics Perception and visualize the intermediate predictive representations through one-step and full step predictions. 

   \item \textbf{F \;\; Qualitative Analysis and Results.} \\
    This section presents comprehensive experiments on simulated and real-world tasks to evaluate the TriVLA framework.
    The simulated experiments employ three benchmarks that encompass diverse robot embodiments and manipulation tasks. In parallel, real-world trials evaluate long-horizon tabletop manipulation using a Kinova Gen3 robotic arm.

   \item \textbf{G \;\; Real-world Experiments}\\
    In this section, we present a series of real-world experiments designed to rigorously evaluate the practical applicability, task generalization, and operational robustness of our TriVLA framework under realistic and unstructured environments.

\end{itemize}

\section{Use of Large Language Models (LLMs)}
In preparing this paper, we used ChatGPT-4o (OpenAI) as a general-purpose writing assistance tool. Its role was strictly limited to checking and improving spelling, grammar, and sentence-level clarity. The LLM did not contribute to the conception of the research idea, experimental design, data analysis, interpretation of results, or the drafting of any substantive scientific content. All intellectual contributions, arguments, and conclusions presented in this paper are our own.

\section{Implementation details}
\label{repro}
\subsection{Training Details.}
As detailed in Section~\ref{method}, we employ a unified triple-system architecture. System 2, the Episodic Multimodal Perception, utilizes the pretrained Eagle-2 VLM for processing visual and language inputs on an NVIDIA H100 GPU. System 3 fine-tunes a video foundation model for manipulation-centric Episodic Dynamics Perception using 193,690 human~\cite{goyal2017something} and 179,074 robotic~\cite{o2023open} trajectories, supplemented by videos from CALVIN ABC, MetaWorld, and real-world tasks. To mitigate dataset discrepancies, we adopt dataset-specific sampling ratios following Octo. Video model fine-tuning requires 2–3 days on 8 NVIDIA H100 GPUs. The generalist policy is subsequently trained on task datasets, taking 5–9 hours on four H100 GPUs.

\subsection{Roll-out Details.} 
The System 2 Episodic Multimodal Perception module employs a pretrained Eagle-2 VLM to extract vision-language tokens, operating at 36.36 Hz on an NVIDIA H100 GPU. In contrast to prior methods, which utilize computationally intensive video denoising—resulting in low control frequencies\cite{black2023zero} or open-loop limitations\cite{du2024learning}—our approach processes each observation through System 2 only once, during the initial forward pass of the Episodic Dynamics Perception module, with inference latency below 85.9 ms. Subsequently, the downstream policy generates a 10-step action chunk~\cite{chi2023diffusion}, enabling control frequencies of 34–36 Hz on a consumer-grade NVIDIA RTX H100 GPU.

\section{Demo Video}
The attached video demonstrates the application of TriVLA, a triple-system architecture inspired by cognitive neuroscience, designed to enhance the capabilities of embodied agents in complex tasks. TriVLA integrates an episodic world model, enabling robots to accumulate, recall, and predict sequential multimodal experiences. This model provides the foundation for robust, adaptive control by simulating episodic memory processes. This showcases the generalization ability of the TriVLA framework. The tasks in the video highlight how TriVLA’s high-level reasoning and dynamic prediction enable robots to handle long-horizon manipulation and understand complex prompts, demonstrating its capability for sophisticated, adaptable decision-making.

The following is a detailed explanation of the tasks that TriVLA handles in the video.

\subsection{Visualization of Episodic Dynamics Perception}
This demo presents the Episodic Dynamics Perception capability of TriVLA in real-world scenarios. By utilizing a stable video diffusion model, TriVLA encodes the current scene and anticipates future dynamics over long time horizons. The visualization highlights ground-truth outcomes, single-step predictions, and full-sequence forecasts, demonstrating how the model captures essential motion cues such as object interactions and robotic arm trajectories. These results show that TriVLA can effectively model entire video sequences and predict future states based on current observations and task instructions, enabling a deeper understanding of physical dynamics for downstream decision-making and action planning.

\subsection{Visualization of Action Trajectory in Simulation}
This demo illustrates the action trajectory generation of TriVLA in a simulated environment. We showcase three representative examples of action sequences executed under multiple consecutive instructions, such as “Pull the handle to open the drawer,” “Grasp and lift the pink block,” “Use the switch to turn on the light bulb,” and “Store the grasped block in the sliding cabin.”
TriVLA demonstrates its ability to comprehend complex instructions, infer underlying intent, and leverage predictive modeling to plan and execute long-horizon tasks. By combining high-level reasoning from vision-language models (VLMs) with dynamic predictive representations from video diffusion models (VDMs), TriVLA integrates world knowledge to enhance intent understanding and predicts future states to guide sequential decision-making. These results highlight how TriVLA effectively coordinates perception, reasoning, and prediction to accomplish complex, multi-step tasks in simulation.

\subsection{Visualization of Action Trajectory in Short-horizon Real-world Tasks}
Across the four short-horizon tasks—folding a pink towel, grasping an orange and placing it on a purple plate, pouring water from a green cup into a purple cup, and relocating a cup to the right side—the TriVLA consistently demonstrates its capability for precise, reliable, and context-aware manipulation. These tasks collectively highlight the model’s versatility: from handling deformable objects to executing accurate pick-and-place operations and controlling dynamic pouring actions. By integrating real-time visual perception with adaptive motor planning, TriVLA achieves robust short-horizon performance, ensuring accurate execution under diverse manipulation challenges. This suite of tasks underscores TriVLA’s efficiency in short-term control, where rapid perception-action coupling is critical for success.

\noindent\textbf{Short-horizon Task: ``Fold Towel"}
In this scenario, the TriVLA demonstrates its capacity to manipulate deformable objects by folding a pink towel with precision. This task requires careful handling of soft materials, demanding spatial reasoning beyond rigid object grasping. The policy leverages its perception to recognize the towel’s shape and orientation, planning an appropriate folding trajectory. Through this task, TriVLA highlights its short-horizon ability to interact with deformable objects in a controlled manner.
\begin{itemize}
    \item TriVLA achieves this by adjusting its grasp and fold strategy in real-time, ensuring the towel is folded along the intended line.
    \item The success of this task emphasizes TriVLA’s adaptability in dealing with non-rigid objects, integrating visual cues into consistent action sequences.
\end{itemize}

\noindent\textbf{Short-horizon Task: ``Grasp Orange"}
In this scenario, the TriVLA showcases its skill in precise object manipulation by picking up an orange and placing it onto a purple plate. The task requires accurate localization and trajectory control, as the system must not only grasp the fruit securely but also transport it safely to the designated location. Through this task, TriVLA demonstrates its ability to carry out reliable short-horizon pick-and-place operations.
\begin{itemize}
    \item TriVLA accomplishes this by dynamically refining its grasp and movement trajectory based on real-time feedback.
    \item The successful completion highlights TriVLA’s integration of perception and motion planning, enabling robust execution of targeted placement tasks.
\end{itemize}

\noindent\textbf{Short-horizon Task: ``Pouring"}
In this scenario, the TriVLA demonstrates its ability to perform liquid transfer by grasping a green cup and pouring water into a purple cup. This task requires stable control of orientation and precise alignment between the two containers, ensuring minimal spillage. The policy leverages its multimodal perception to model both rigid object states and the flow dynamics of the liquid. Through this task, TriVLA demonstrates competence in controlled pouring, a challenging short-horizon manipulation.
\begin{itemize}
    \item TriVLA achieves this by continuously monitoring the cup angle and relative position to regulate water flow.
    \item The success of this task highlights TriVLA’s ability to coordinate fine-grained motor control with visual guidance for dynamic object interaction.
\end{itemize}

\noindent\textbf{Short-horizon Task: ``Pick Cup"}
In this scenario, the TriVLA performs a straightforward relocation task by picking up a cup from the table and placing it on the right side. While simple, this task demands accurate detection of the cup’s position and a reliable transfer motion without disrupting the environment. Through this task, TriVLA showcases its efficiency in executing basic, short-horizon manipulation.
\begin{itemize}
    \item TriVLA achieves this by generating a direct grasp-to-place trajectory, adapting its motion based on sensory feedback.
    \item The success of this task demonstrates TriVLA’s ability to reliably execute fundamental pick-and-place operations with consistency.
\end{itemize}

\subsection{Visualization of Action Trajectory in long-horizon Real-world Tasks}

\noindent\textbf{Long-horizon Task: ``Beverage Preparation”}
In this scenario, the TriVLA demonstrates its competence in executing a complex, sequential task: picking up a red beverage can from the shelf, pouring the beverage into a yellow cup, inserting a straw into the yellow cup, and finally pushing the cup forward. Unlike short-horizon manipulations, this task requires the integration of multiple atomic actions into a coherent sequence, demanding sustained spatial reasoning, memory of intermediate states, and precise coordination across distinct phases. Through this task, TriVLA showcases its ability to plan and execute long-horizon activities where success depends on maintaining consistency across multiple dependent steps.
\begin{itemize}
    \item TriVLA achieves this by decomposing the task into sub-goals, dynamically adjusting its strategy based on real-time perception and the evolving state of the environment.
    \item The successful completion highlights TriVLA’s ability to combine high-level planning with fine-grained control, ensuring the transition is seamless and reliable.
    \item This task demonstrates TriVLA’s strength in long-horizon reasoning, where sustained action sequences and contextual understanding are essential for achieving complex goals.
\end{itemize}

TriVLA’s episodic world model enables robots to simulate memory processes, allowing them to accumulate sequential experiences and predict future actions. This capability helps robots adapt dynamically to changing environments and illustrates how embodied agents can reason about actions and experiences in a human-like way. It adopts a triple-system architecture that integrates episodic memory, high-level reasoning, and dynamic prediction. This unified design allows robots to understand multi-step tasks, solve complex manipulation problems, and make decisions grounded in both past experiences and anticipated outcomes.

In summary, TriVLA provides a robust framework for robots, offering spatial-temporal awareness, high-level reasoning, and adaptive control over long horizons. The model demonstrates exceptional generalization ability, enabling robots to perform tasks in diverse, complex environments.

\section{Comparison Methods}
Generalist robot policies have been extensively investigated in prior research.
In our experiments, we select a representative subset of prior methods for comparison, focusing on those that have achieved state-of-the-art performance or employ approaches similar to ours.
\begin{itemize}
    \item RT-1~\cite{brohan2022rt}: A general action learning robot policy integrating semantic features via Efficient-Net with FiLM-conditioning, subsequently employing token learners.
    \item Diffusion Policy~\cite{chi2023diffusion}: A action learning approach modeling the robot’s visuomotor policy as a conditional denoising diffusion process enhanced with action diffusers.
    \item Robo-Flamingo~\cite{li2023vision}: A direct action learning policy leveraging a pre-trained LLM, integrating visual information into each layer following the Flamingo~\cite{alayrac2022flamingo}.
    \item UniPi~\cite{du2024learning}: Initiates by training a video prediction model for future sequence generation and concurrently learns an inverse kinematics model between frames to infer actions.
    \item MDT~\cite{reuss2024multimodaldiffusiontransformerlearning}: Trains a diffusion transformer-based policy complemented by an auxiliary MAE loss to facilitate future state reconstruction.
    \item Susie~\cite{black2023zero}: Employs a fine-tuned InstructPix2Pix~\cite{brooks2023instructpix2pix} model to generate goal images and trains a downstream diffusion policy conditioned on these goal images.
    \item GR-1~\cite{wu2023unleashing}: Models video and action sequences using an autoregressive transformer. During policy execution, GR-1 predicts one future frame followed by a action.
    \item Robo-Uniview~\cite{liu2024robouniview}: Develops a 3D-aware visual encoder supervised by a 3D occupancy loss for policy learning.
    \item Vidman~\cite{wen2024vidman}: Pre-trained on the Open X-Embodiment video dataset, it employs a self-attention adapter to convert video representations into policies. However, Vidman’s performance is suboptimal due to the absence of fine-tuning the video model on downstream tasks.
    \item Seer~\cite{tian2024predictive}: Designs a novel end-to-end framework that leverages predictive inverse dynamics models to integrate vision and action for scalable robotic manipulation. 
    \item VPP~\cite{hu2024video}: Leverages video diffusion models to generate visual representations, addressing the limitations of traditional vision encoders in capturing temporal aspects critical for robotic manipulation. 
\end{itemize}

\begin{figure*}[t]
	\centering
	\includegraphics[width=\linewidth]{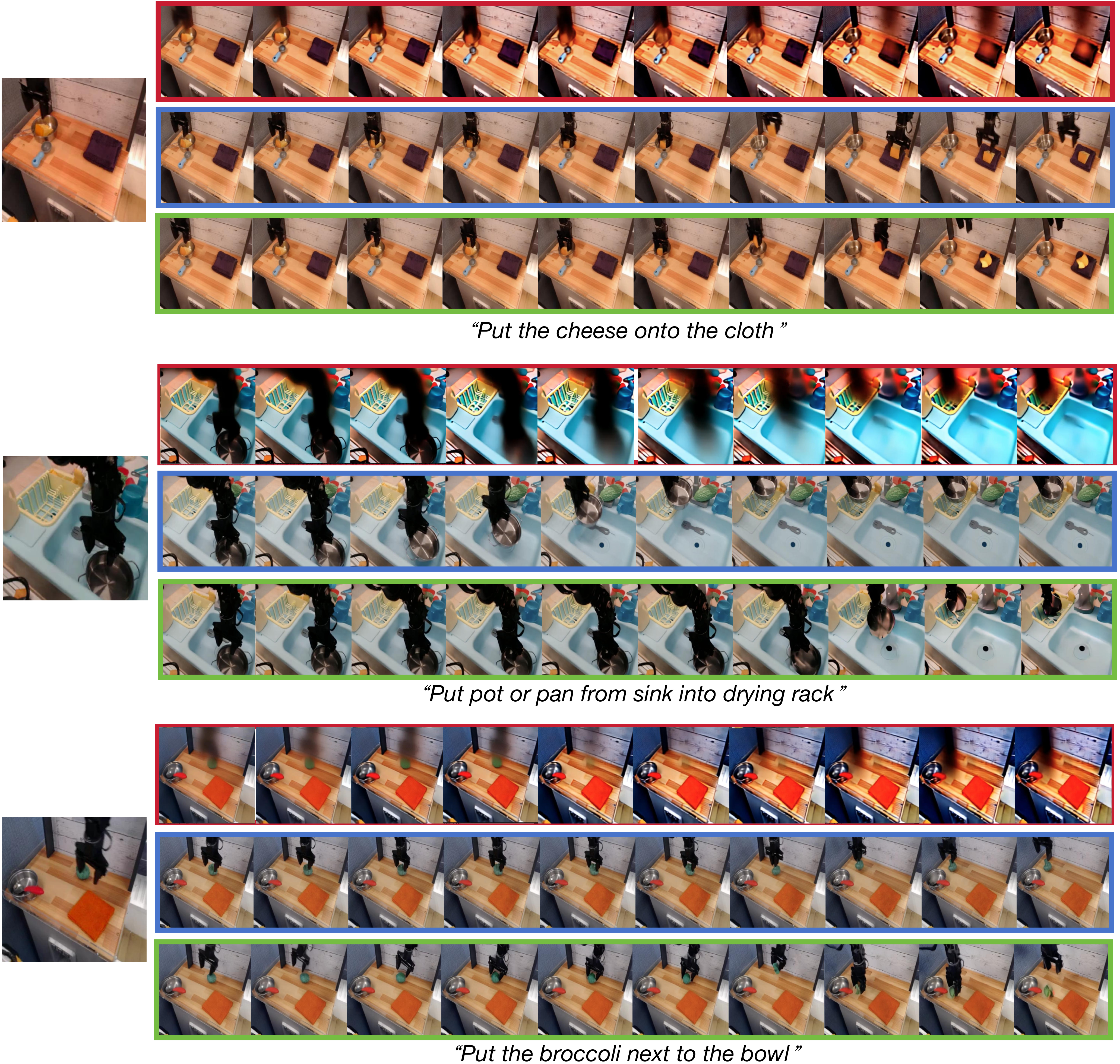}
    \vspace{-0.2in}
	\caption{{\textbf{Visualization of Episodic Dynamics Perception on the Bridge Benchmark.} 
    The red box indicates one-step prediction, the blue box corresponds to full-step prediction, and the green box marks the ground truth. We can observe that the representation can provide valuable information on physical dynamics, although
the textures and details are not precise enough.	
    \label{vp_supp}}}
    \vspace{-0.1in}
\end{figure*}

\section{Details and More Results of Episodic Dynamics Perception}
We employ a stable video diffusion model as the dynamics perception module, performing a single forward pass to obtain visual representations that
encompass both current static information and predicted future dynamics. As illustrated in Figure~\ref{vp_supp},
we present visualizations of ground-truth futures alongside single-step and full-step predictions on the Bridge benchmark. The visualization results indicate that single-step representations
convey critical information, including object and robot arm motion, thereby effectively supporting downstream action learning. The dynamics perception module models entire video sequences and predicts future frames conditioned on current observations and instructions, demonstrating a sound
understanding of physical dynamics.

\begin{figure*}[t]
	\centering
	\includegraphics[width=\linewidth]{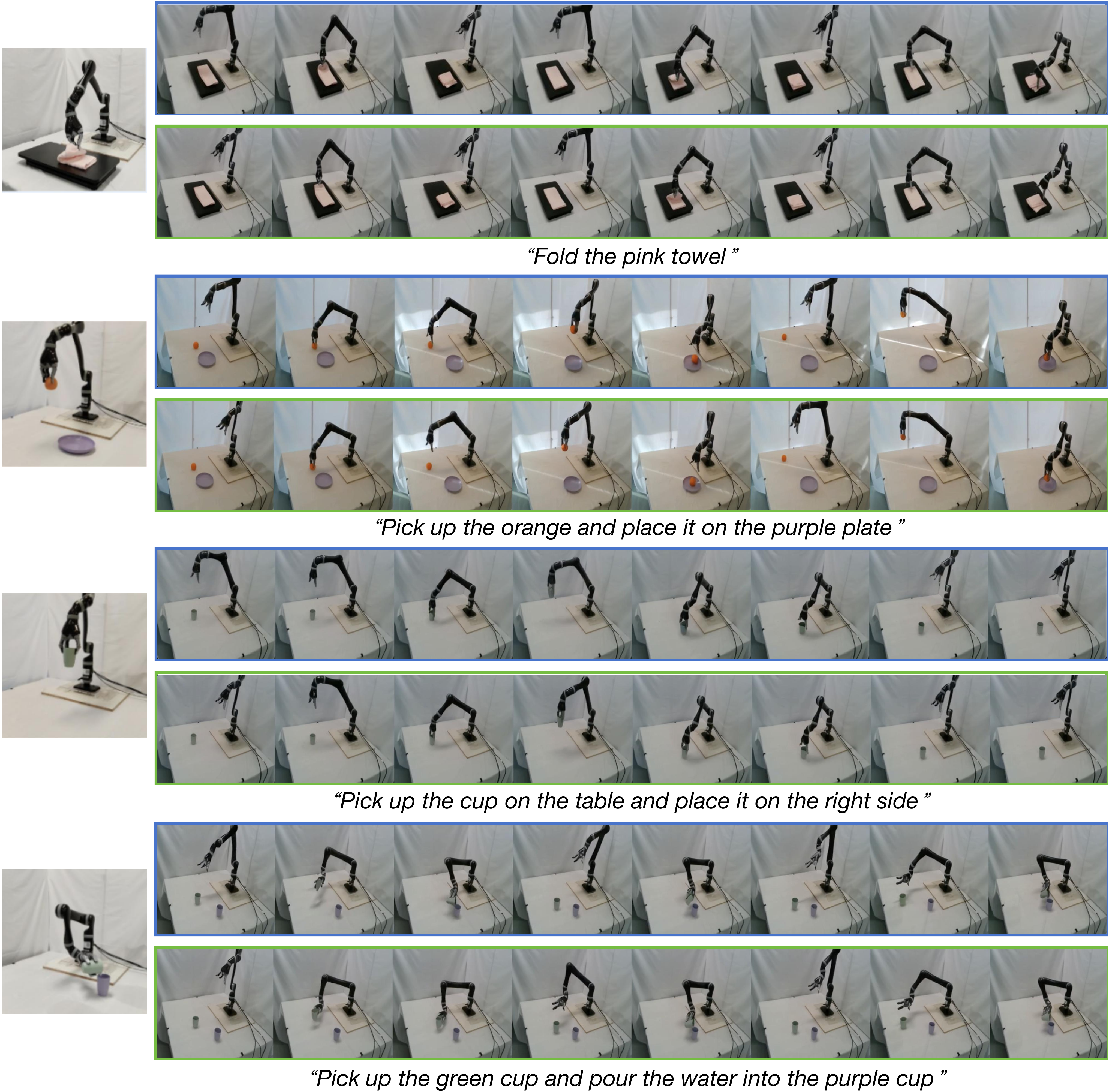}
    \vspace{-0.2in}
	\caption{{\textbf{Visualization of Episodic Dynamics Perception on Real-world Tasks.} 
    The blue box corresponds to the full-step prediction, and the green box marks the ground truth of the current timestep. We can observe that representation can provide valuable information on physical dynamics, although
the textures and details are not precise enough.	
    \label{vp_supp_real}}}
    \vspace{-0.1in}
\end{figure*}

\begin{figure*}[t]
	\centering
	\includegraphics[width=\linewidth]{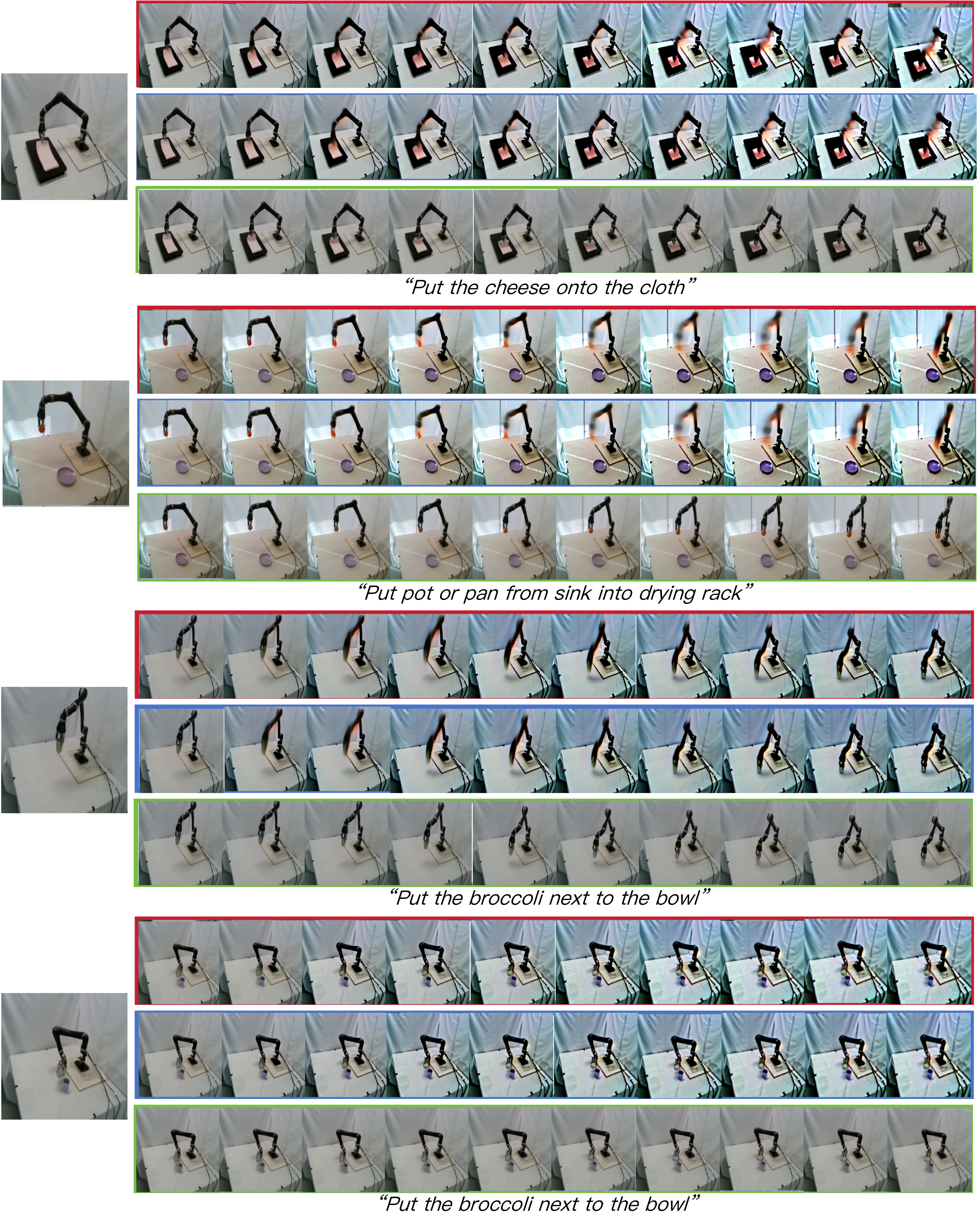}
    \vspace{-0.2in}
	\caption{{\textbf{Visualization of Episodic Dynamics Perception on Real-world Tasks.} 
    The red box indicates the one-step prediction, the blue box corresponds to the full-step prediction, and the green box marks the ground truth. We can observe that representation can provide valuable information on physical dynamics, although
the textures and details are not precise enough.	
    \label{vp_supp_real_3}}}
    \vspace{-0.1in}
\end{figure*}

Additionally, we extend our analysis to real-world experiments, where we replicate the same predictive framework on an actual robot. As shown in Figure~\ref{vp_supp_real} and ~\ref{vp_supp_real_3}, the true-to-life experimental results mirror the findings from the simulated setup, further validating the robustness of our prediction model. These real-world predictions are crucial, as they show that our system not only captures the motion of objects and the robot arm but also adapts to real-world uncertainties, such as sensor noise and minor mechanical inaccuracies.

The real-world prediction visualizations highlight several important aspects of our approach:
\begin{itemize}
    \item Accurate Motion Forecasting: Even in the face of real-world complexities, the system accurately predicts future motions of both the robot arm and surrounding objects. This is key for enabling high-level decision-making and adaptive action execution.
    \item Real-World Generalization: The model demonstrates strong generalization capabilities, transferring learned predictions from the benchmark environment to practical settings without requiring extensive retraining. The system's robustness in handling real-world dynamics proves the versatility of the proposed architecture.
\end{itemize}

By leveraging these visualizations and predictions in both the simulated and real environments, we show that our framework can bridge the gap between theoretical modeling and real-world robotic applications. This provides a powerful tool for task generalization, enabling robots to efficiently plan and execute complex actions in diverse scenarios.
In summary, the combination of simulated and real-world results not only validates the robustness of our prediction framework but also underscores its potential for real-time action learning and autonomous decision-making in physical environments. The visualizations of future predictions further support the importance of incorporating dynamic modeling into robotic systems, fostering a deeper understanding of physical interactions and improving the overall system performance.

\begin{figure*}[t]
	\centering
	\includegraphics[width=\linewidth]{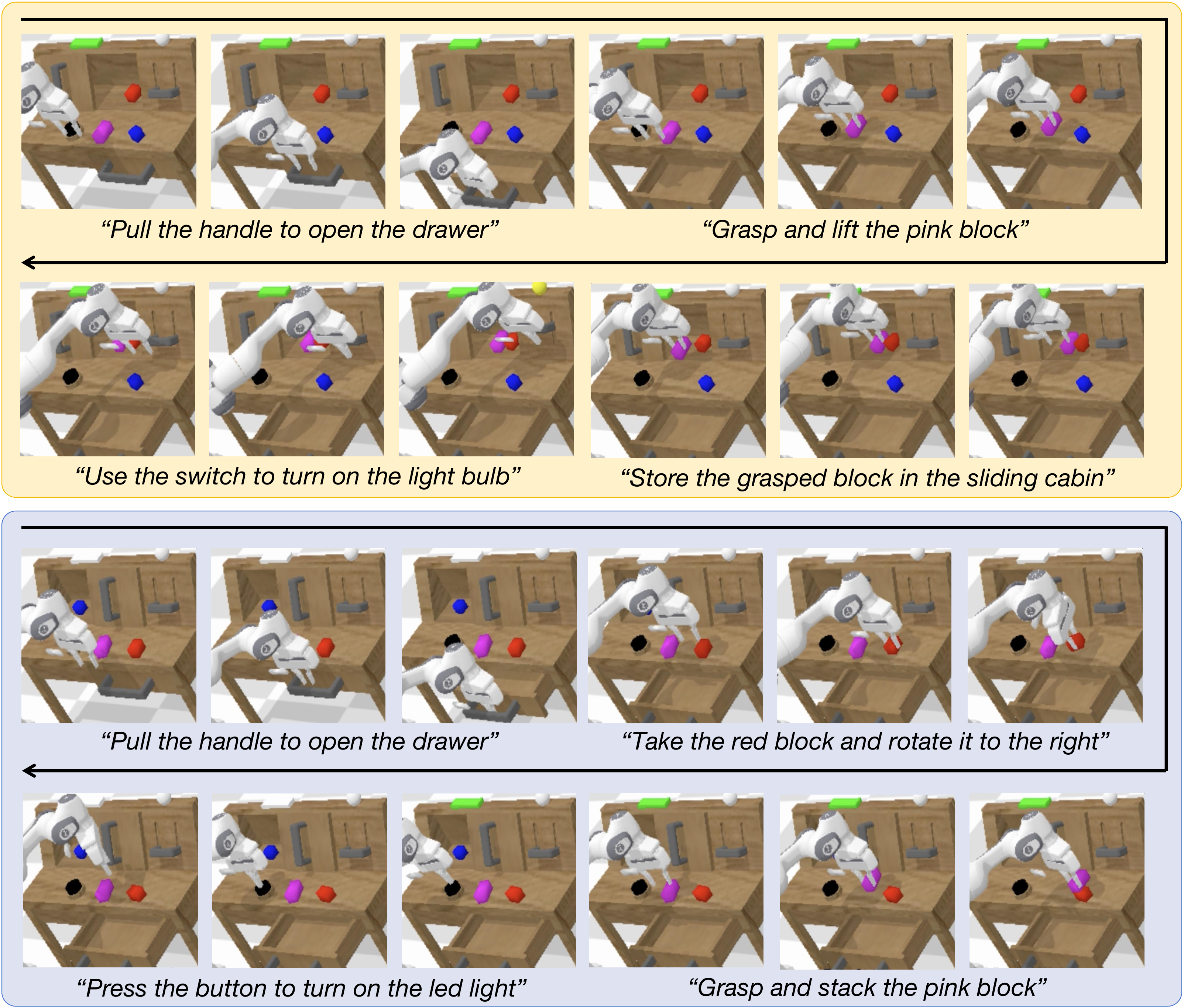}
    \vspace{-0.2in}
	\caption{\textbf{Qualitative case study of CALVIN benchmark.} Our \textbf{TriVLA} performs strongly in long-horizon missions. For example, in the CALVIN simulation task, it integrates world knowledge to interpret intent and uses a world model to predict future states. Given multiple sequential instructions, TriVLA can effectively execute long-horizon tasks.
	\label{qual_1_supp}}
    \vspace{-3mm}
\end{figure*}
\section{Qualitative Analysis and Results}
We provide qualitative examples of action sequences generated
by TriVLA in Figure~\ref{qual_1_supp}. Given multiple consecutive instructions—for instance, “Pull the handle to
open the drawer,” “Grasp and lift the pink block,” “Use the switch to turn on the light bulb,” and
“Store the grasped block in the sliding cabin”—TriVLA demonstrates the ability to comprehend
instructions, infer intent, and utilize predictive capabilities to accomplish long-horizon tasks. The results demonstrate that TriVLA employs VLMs and VDMs for both high-level reasoning based on
common knowledge and dynamic predictive representation provided by a world model. TriVLA
integrates world knowledge to enhance intent understanding and utilizes a world model for future
state prediction when processing multiple sequential instructions, thereby enabling effective long horizon task execution.
\begin{itemize}
    \item Precise Alignment with Modeled Dynamics: Since simulation provides deterministic or near-deterministic dynamics, TriVLA demonstrates highly consistent outcomes across repeated trials. For instance, during multi-step manipulation sequences, object trajectories match predicted states almost perfectly, showcasing the model’s capability to exploit stable environments for accurate planning.  
    \item Stress Testing under Controlled Perturbations: Simulation allows for the systematic injection of domain variations, such as randomized object masses, altered friction coefficients, or unexpected collisions. TriVLA adapts to these controlled perturbations by updating its predictions accordingly, highlighting its resilience under a wide spectrum of simulated uncertainties.  
    \item Long-Horizon Reasoning at Scale: Most importantly, simulation provides an ideal platform for TriVLA to validate long-horizon planning strategies across diverse scenarios. By anticipating future states over extended horizons, the model learns to optimize its policy in a wide variety of contexts, generating transferable skills that can later be fine-tuned for real-world execution.  
\end{itemize}

\section{Real-world experiments}
\begin{figure*}[t]
	\centering
	\includegraphics[width=\linewidth]{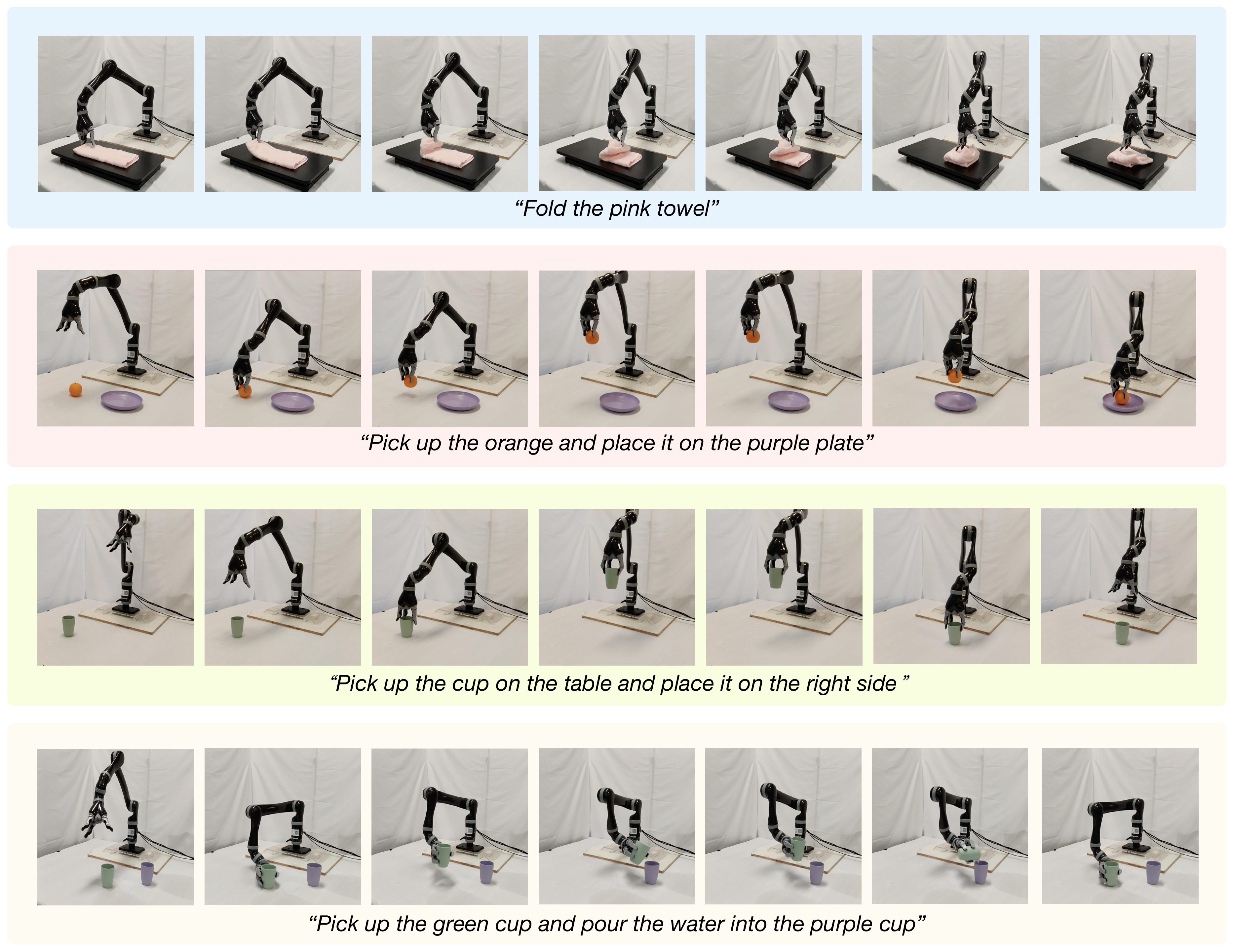}
    \vspace{-0.2in}
	\caption{\textbf{Qualitative case study of real-world tasks.} Our \textbf{TriVLA} performs well in real-world tasks, successfully executing both short-horizon and long-horizon manipulations. The results illustrate its ability to integrate perception, prediction, and control for reliable task completion under real-world conditions.
	\label{qual_2_supp}}
    \vspace{-3mm}
\end{figure*}

In addition to the simulated results, we also conducted real-world experiments, where TriVLA successfully generates and executes action sequences on a physical robot. As shown in Figure~\ref{qual_2_supp}, the robot accurately follows the same sequence of tasks, starting from pulling the handle to opening the drawer, grasping and lifting the pink block, using the switch to turn on the light bulb, and finally storing the block in the sliding cabin. These real-world results closely align with the predictions and actions generated in the simulated environment, further validating TriVLA's ability to handle complex, multi-step tasks in dynamic, real-world scenarios.

The real-world experiments highlight several key advantages of TriVLA’s approach:
\begin{itemize}
    \item Real-Time Sequence Execution: The robot efficiently processes and executes long-horizon tasks in real-time, leveraging TriVLA's ability to predict intermediate states and adjust actions accordingly. Despite the inherent variability and unpredictability of the real world, such as slight environmental changes or sensor noise, TriVLA’s predictive capabilities allow the robot to remain on task without requiring extensive retraining.
    \item High Fidelity in Task Completion: As the robot progresses through the sequence of actions, it demonstrates a strong alignment between predicted outcomes and actual results. For instance, after pulling the handle, the drawer opens correctly, and the robot adjusts its grip on the block while maintaining stability during the lift. This showcases the robustness of TriVLA’s predictive world model in real-world settings.
    \item Dynamic Adaptation to Uncertainty: The real-world setup also presents challenges like minor inaccuracies in motor control or shifting environmental conditions. TriVLA exhibits impressive adaptability, dynamically adjusting predictions and actions to account for these uncertainties, ensuring continued task success.
    \item Long-Horizon Task Planning: Perhaps most notably, TriVLA demonstrates its ability to execute long-horizon plans by integrating both episodic memory and predictive reasoning. By leveraging its world model, TriVLA is able to anticipate future states and proactively adjust actions, ensuring that all steps of the sequence are successfully completed, even in the presence of unforeseen challenges.
\end{itemize}

Overall, these real-world experiments reinforce the core strength of TriVLA: its ability to understand complex instructions, reason about sequential actions, and predict future states—an essential capability for enabling embodied agents to perform sophisticated tasks autonomously and effectively in the real world.
Through the combination of simulated and real-world action sequence generation, TriVLA proves to be a highly capable architecture for long-horizon task execution, paving the way for more advanced and adaptable autonomous systems.

\end{document}

%% file: math_commands.tex
\usepackage{amsmath,amsfonts,bm}

\def\eqref#1{equation~\ref{#1}}

\def\1{\bm{1}}

\DeclareMathAlphabet{\mathsfit}{\encodingdefault}{\sfdefault}{m}{sl}
\SetMathAlphabet{\mathsfit}{bold}{\encodingdefault}{\sfdefault}{bx}{n}













%% file: iclr2026_conference.bbl
\begin{thebibliography}{67}
\providecommand{\natexlab}[1]{#1}
\providecommand{\url}[1]{\texttt{#1}}
\expandafter\ifx\csname urlstyle\endcsname\relax
  \providecommand{\doi}[1]{doi: #1}\else
  \providecommand{\doi}{doi: \begingroup \urlstyle{rm}\Url}\fi

\bibitem[Ahn et~al.(2022)Ahn, Brohan, Brown, Chebotar, Cortes, David, Finn, Fu, Gopalakrishnan, Hausman, et~al.]{ahn2022can}
Michael Ahn, Anthony Brohan, Noah Brown, Yevgen Chebotar, Omar Cortes, Byron David, Chelsea Finn, Chuyuan Fu, Keerthana Gopalakrishnan, Karol Hausman, et~al.
\newblock Do as i can, not as i say: Grounding language in robotic affordances.
\newblock \emph{arXiv preprint arXiv:2204.01691}, 2022.

\bibitem[Alayrac et~al.(2022)Alayrac, Donahue, Luc, Miech, Barr, Hasson, Lenc, Mensch, Millican, Reynolds, et~al.]{alayrac2022flamingo}
Jean-Baptiste Alayrac, Jeff Donahue, Pauline Luc, Antoine Miech, Iain Barr, Yana Hasson, Karel Lenc, Arthur Mensch, Katherine Millican, Malcolm Reynolds, et~al.
\newblock Flamingo: a visual language model for few-shot learning.
\newblock \emph{Advances in Neural Information Processing Systems}, 35:\penalty0 23716--23736, 2022.

\bibitem[Bai et~al.(2023)Bai, Bai, Yang, Wang, Tan, Wang, Lin, Zhou, and Zhou]{bai2023qwen}
Jinze Bai, Shuai Bai, Shusheng Yang, Shijie Wang, Sinan Tan, Peng Wang, Junyang Lin, Chang Zhou, and Jingren Zhou.
\newblock Qwen-vl: A frontier large vision-language model with versatile abilities.
\newblock \emph{arXiv preprint arXiv:2308.12966}, 2023.

\bibitem[Belkhale et~al.(2024)Belkhale, Ding, Xiao, Sermanet, Vuong, Tompson, Chebotar, Dwibedi, and Sadigh]{belkhale2024rt}
Suneel Belkhale, Tianli Ding, Ted Xiao, Pierre Sermanet, Quon Vuong, Jonathan Tompson, Yevgen Chebotar, Debidatta Dwibedi, and Dorsa Sadigh.
\newblock Rt-h: Action hierarchies using language.
\newblock \emph{arXiv preprint arXiv:2403.01823}, 2024.

\bibitem[Bharadhwaj et~al.(2024)Bharadhwaj, Dwibedi, Gupta, Tulsiani, Doersch, Xiao, Shah, Xia, Sadigh, and Kirmani]{bharadhwaj2024gen2act}
Homanga Bharadhwaj, Debidatta Dwibedi, Abhinav Gupta, Shubham Tulsiani, Carl Doersch, Ted Xiao, Dhruv Shah, Fei Xia, Dorsa Sadigh, and Sean Kirmani.
\newblock Gen2act: Human video generation in novel scenarios enables generalizable robot manipulation.
\newblock \emph{arXiv preprint arXiv:2409.16283}, 2024.

\bibitem[Bjorck et~al.(2025)Bjorck, Casta{\~n}eda, Cherniadev, Da, Ding, Fan, Fang, Fox, Hu, Huang, et~al.]{bjorck2025gr00t}
Johan Bjorck, Fernando Casta{\~n}eda, Nikita Cherniadev, Xingye Da, Runyu Ding, Linxi Fan, Yu~Fang, Dieter Fox, Fengyuan Hu, Spencer Huang, et~al.
\newblock Gr00t n1: An open foundation model for generalist humanoid robots.
\newblock \emph{arXiv preprint arXiv:2503.14734}, 2025.

\bibitem[Black et~al.(2023)Black, Nakamoto, Atreya, Walke, Finn, Kumar, and Levine]{black2023zero}
Kevin Black, Mitsuhiko Nakamoto, Pranav Atreya, Homer Walke, Chelsea Finn, Aviral Kumar, and Sergey Levine.
\newblock Zero-shot robotic manipulation with pretrained image-editing diffusion models.
\newblock \emph{arXiv preprint arXiv:2310.10639}, 2023.

\bibitem[Black et~al.(2024)Black, Brown, Driess, Esmail, Equi, Finn, Fusai, Groom, Hausman, Ichter, Jakubczak, Jones, Ke, Levine, Li-Bell, Mothukuri, Nair, Pertsch, Shi, Tanner, Vuong, Walling, Wang, and Zhilinsky]{black2024pi0visionlanguageactionflowmodel}
Kevin Black, Noah Brown, Danny Driess, Adnan Esmail, Michael Equi, Chelsea Finn, Niccolo Fusai, Lachy Groom, Karol Hausman, Brian Ichter, Szymon Jakubczak, Tim Jones, Liyiming Ke, Sergey Levine, Adrian Li-Bell, Mohith Mothukuri, Suraj Nair, Karl Pertsch, Lucy~Xiaoyang Shi, James Tanner, Quan Vuong, Anna Walling, Haohuan Wang, and Ury Zhilinsky.
\newblock $\pi_0$: A vision-language-action flow model for general robot control, 2024.
\newblock URL \url{https://arxiv.org/abs/2410.24164}.

\bibitem[Blattmann et~al.(2023{\natexlab{a}})Blattmann, Dockhorn, Kulal, Mendelevitch, Kilian, Lorenz, Levi, English, Voleti, Letts, et~al.]{blattmann2023stable}
Andreas Blattmann, Tim Dockhorn, Sumith Kulal, Daniel Mendelevitch, Maciej Kilian, Dominik Lorenz, Yam Levi, Zion English, Vikram Voleti, Adam Letts, et~al.
\newblock Stable video diffusion: Scaling latent video diffusion models to large datasets.
\newblock \emph{arXiv preprint arXiv:2311.15127}, 2023{\natexlab{a}}.

\bibitem[Blattmann et~al.(2023{\natexlab{b}})Blattmann, Rombach, Ling, Dockhorn, Kim, Fidler, and Kreis]{blattmann2023align}
Andreas Blattmann, Robin Rombach, Huan Ling, Tim Dockhorn, Seung~Wook Kim, Sanja Fidler, and Karsten Kreis.
\newblock Align your latents: High-resolution video synthesis with latent diffusion models.
\newblock In \emph{Proceedings of the IEEE/CVF Conference on Computer Vision and Pattern Recognition}, pp.\  22563--22575, 2023{\natexlab{b}}.

\bibitem[Blundell et~al.(2016)Blundell, Uria, Pritzel, Li, Ruderman, Leibo, Rae, Wierstra, and Hassabis]{blundell2016model}
Charles Blundell, Benigno Uria, Alexander Pritzel, Yazhe Li, Avraham Ruderman, Joel~Z Leibo, Jack Rae, Daan Wierstra, and Demis Hassabis.
\newblock Model-free episodic control.
\newblock \emph{arXiv preprint arXiv:1606.04460}, 2016.

\bibitem[Brohan et~al.(2022)Brohan, Brown, Carbajal, Chebotar, Dabis, Finn, Gopalakrishnan, Hausman, Herzog, Hsu, et~al.]{brohan2022rt}
Anthony Brohan, Noah Brown, Justice Carbajal, Yevgen Chebotar, Joseph Dabis, Chelsea Finn, Keerthana Gopalakrishnan, Karol Hausman, Alex Herzog, Jasmine Hsu, et~al.
\newblock Rt-1: Robotics transformer for real-world control at scale.
\newblock \emph{arXiv preprint arXiv:2212.06817}, 2022.

\bibitem[Brohan et~al.(2023)Brohan, Brown, Carbajal, Chebotar, Chen, Choromanski, Ding, Driess, Dubey, Finn, et~al.]{brohan2023rt}
Anthony Brohan, Noah Brown, Justice Carbajal, Yevgen Chebotar, Xi~Chen, Krzysztof Choromanski, Tianli Ding, Danny Driess, Avinava Dubey, Chelsea Finn, et~al.
\newblock Rt-2: Vision-language-action models transfer web knowledge to robotic control.
\newblock \emph{arXiv preprint arXiv:2307.15818}, 2023.

\bibitem[Brooks et~al.(2023)Brooks, Holynski, and Efros]{brooks2023instructpix2pix}
Tim Brooks, Aleksander Holynski, and Alexei~A Efros.
\newblock Instructpix2pix: Learning to follow image editing instructions.
\newblock In \emph{Proceedings of the IEEE/CVF Conference on Computer Vision and Pattern Recognition}, pp.\  18392--18402, 2023.

\bibitem[Brooks et~al.(2024)Brooks, Peebles, Holmes, DePue, Guo, Jing, Schnurr, Taylor, Luhman, Luhman, Ng, Wang, and Ramesh]{videoworldsimulators2024}
Tim Brooks, Bill Peebles, Connor Holmes, Will DePue, Yufei Guo, Li~Jing, David Schnurr, Joe Taylor, Troy Luhman, Eric Luhman, Clarence Ng, Ricky Wang, and Aditya Ramesh.
\newblock Video generation models as world simulators.
\newblock 2024.
\newblock URL \url{https://openai.com/research/video-generation-models-as-world-simulators}.

\bibitem[Chen et~al.(2024)Chen, Guo, He, Zhang, Zhang, Yang, Zhao, and Bian]{chen2024igor}
Xiaoyu Chen, Junliang Guo, Tianyu He, Chuheng Zhang, Pushi Zhang, Derek~Cathera Yang, Li~Zhao, and Jiang Bian.
\newblock Igor: Image-goal representations are the atomic control units for foundation models in embodied ai.
\newblock \emph{arXiv preprint arXiv:2411.00785}, 2024.

\bibitem[Chi et~al.(2023)Chi, Xu, Feng, Cousineau, Du, Burchfiel, Tedrake, and Song]{chi2023diffusion}
Cheng Chi, Zhenjia Xu, Siyuan Feng, Eric Cousineau, Yilun Du, Benjamin Burchfiel, Russ Tedrake, and Shuran Song.
\newblock Diffusion policy: Visuomotor policy learning via action diffusion.
\newblock \emph{The International Journal of Robotics Research}, pp.\  02783649241273668, 2023.

\bibitem[Driess et~al.(2023)Driess, Xia, Sajjadi, Lynch, Chowdhery, Ichter, Wahid, Tompson, Vuong, Yu, et~al.]{driess2023palm}
Danny Driess, Fei Xia, Mehdi~SM Sajjadi, Corey Lynch, Aakanksha Chowdhery, Brian Ichter, Ayzaan Wahid, Jonathan Tompson, Quan Vuong, Tianhe Yu, et~al.
\newblock Palm-e: An embodied multimodal language model.
\newblock \emph{arXiv preprint arXiv:2303.03378}, 2023.

\bibitem[Du et~al.(2024)Du, Yang, Dai, Dai, Nachum, Tenenbaum, Schuurmans, and Abbeel]{du2024learning}
Yilun Du, Sherry Yang, Bo~Dai, Hanjun Dai, Ofir Nachum, Josh Tenenbaum, Dale Schuurmans, and Pieter Abbeel.
\newblock Learning universal policies via text-guided video generation.
\newblock \emph{Advances in Neural Information Processing Systems}, 36, 2024.

\bibitem[Gao* et~al.(2023)Gao*, Han*, Zhang*, Lin, Geng, Zhou, Zhang, Lu, He, Yue, et~al.]{gao2023llama}
Peng Gao*, Jiaming Han*, Renrui Zhang*, Ziyi Lin, Shijie Geng, Aojun Zhou, Wei Zhang, Pan Lu, Conghui He, Xiangyu Yue, et~al.
\newblock Llama-adapter v2: Parameter-efficient visual instruction model.
\newblock \emph{arXiv preprint arXiv:2304.15010}, 2023.

\bibitem[Gershman \& Daw(2017)Gershman and Daw]{gershman2017reinforcement}
Samuel~J Gershman and Nathaniel~D Daw.
\newblock Reinforcement learning and episodic memory in humans and animals: an integrative framework.
\newblock \emph{Annual review of psychology}, 68\penalty0 (1):\penalty0 101--128, 2017.

\bibitem[Goyal et~al.(2017)Goyal, Ebrahimi~Kahou, Michalski, Materzynska, Westphal, Kim, Haenel, Fruend, Yianilos, Mueller-Freitag, et~al.]{goyal2017something}
Raghav Goyal, Samira Ebrahimi~Kahou, Vincent Michalski, Joanna Materzynska, Susanne Westphal, Heuna Kim, Valentin Haenel, Ingo Fruend, Peter Yianilos, Moritz Mueller-Freitag, et~al.
\newblock The" something something" video database for learning and evaluating visual common sense.
\newblock In \emph{Proceedings of the IEEE international conference on computer vision}, pp.\  5842--5850, 2017.

\bibitem[Graves \& Graves(2012)Graves and Graves]{graves2012long}
Alex Graves and Alex Graves.
\newblock Long short-term memory.
\newblock \emph{Supervised sequence labelling with recurrent neural networks}, pp.\  37--45, 2012.

\bibitem[Guo et~al.(2024)Guo, Hu, Zhang, Wang, Chen, Lu, and Chen]{guo2024prediction}
Yanjiang Guo, Yucheng Hu, Jianke Zhang, Yen-Jen Wang, Xiaoyu Chen, Chaochao Lu, and Jianyu Chen.
\newblock Prediction with action: Visual policy learning via joint denoising process.
\newblock In \emph{The Thirty-eighth Annual Conference on Neural Information Processing Systems}, 2024.

\bibitem[Hong et~al.(2022)Hong, Ding, Zheng, Liu, and Tang]{hong2022cogvideo}
Wenyi Hong, Ming Ding, Wendi Zheng, Xinghan Liu, and Jie Tang.
\newblock Cogvideo: Large-scale pretraining for text-to-video generation via transformers.
\newblock \emph{arXiv preprint arXiv:2205.15868}, 2022.

\bibitem[Hu et~al.(2024)Hu, Guo, Wang, Chen, Wang, Zhang, Sreenath, Lu, and Chen]{hu2024video}
Yucheng Hu, Yanjiang Guo, Pengchao Wang, Xiaoyu Chen, Yen-Jen Wang, Jianke Zhang, Koushil Sreenath, Chaochao Lu, and Jianyu Chen.
\newblock Video prediction policy: A generalist robot policy with predictive visual representations.
\newblock \emph{arXiv preprint arXiv:2412.14803}, 2024.

\bibitem[Huang et~al.(2024{\natexlab{a}})Huang, Ponomarenko, Jiang, Li, Hu, Gao, Li, and Dong]{huang2024manipvqa}
Siyuan Huang, Iaroslav Ponomarenko, Zhengkai Jiang, Xiaoqi Li, Xiaobin Hu, Peng Gao, Hongsheng Li, and Hao Dong.
\newblock Manipvqa: Injecting robotic affordance and physically grounded information into multi-modal large language models.
\newblock \emph{arXiv preprint arXiv:2403.11289}, 2024{\natexlab{a}}.

\bibitem[Huang et~al.(2023)Huang, Wang, Zhang, Li, Wu, and Fei-Fei]{huang2023voxposer}
Wenlong Huang, Chen Wang, Ruohan Zhang, Yunzhu Li, Jiajun Wu, and Li~Fei-Fei.
\newblock Voxposer: Composable 3d value maps for robotic manipulation with language models.
\newblock \emph{arXiv preprint arXiv:2307.05973}, 2023.

\bibitem[Huang et~al.(2024{\natexlab{b}})Huang, Wang, Li, Zhang, and Fei-Fei]{huang2024rekep}
Wenlong Huang, Chen Wang, Yunzhu Li, Ruohan Zhang, and Li~Fei-Fei.
\newblock Rekep: Spatio-temporal reasoning of relational keypoint constraints for robotic manipulation.
\newblock \emph{arXiv preprint arXiv:2409.01652}, 2024{\natexlab{b}}.

\bibitem[Intelligence et~al.(2025)Intelligence, Black, Brown, Darpinian, Dhabalia, Driess, Esmail, Equi, Finn, Fusai, Galliker, Ghosh, Groom, Hausman, Ichter, Jakubczak, Jones, Ke, LeBlanc, Levine, Li-Bell, Mothukuri, Nair, Pertsch, Ren, Shi, Smith, Springenberg, Stachowicz, Tanner, Vuong, Walke, Walling, Wang, Yu, and Zhilinsky]{intelligence2025pi05visionlanguageactionmodelopenworld}
Physical Intelligence, Kevin Black, Noah Brown, James Darpinian, Karan Dhabalia, Danny Driess, Adnan Esmail, Michael Equi, Chelsea Finn, Niccolo Fusai, Manuel~Y. Galliker, Dibya Ghosh, Lachy Groom, Karol Hausman, Brian Ichter, Szymon Jakubczak, Tim Jones, Liyiming Ke, Devin LeBlanc, Sergey Levine, Adrian Li-Bell, Mohith Mothukuri, Suraj Nair, Karl Pertsch, Allen~Z. Ren, Lucy~Xiaoyang Shi, Laura Smith, Jost~Tobias Springenberg, Kyle Stachowicz, James Tanner, Quan Vuong, Homer Walke, Anna Walling, Haohuan Wang, Lili Yu, and Ury Zhilinsky.
\newblock $\pi_{0.5}$: a vision-language-action model with open-world generalization, 2025.
\newblock URL \url{https://arxiv.org/abs/2504.16054}.

\bibitem[Jin et~al.(2024)Jin, Li, Shi, Hao, Sun, Zhang, Fang, et~al.]{jin2024robotgpt}
Yixiang Jin, Dingzhe Li, Jun Shi, Peng Hao, Fuchun Sun, Jianwei Zhang, Bin Fang, et~al.
\newblock Robotgpt: Robot manipulation learning from chatgpt.
\newblock \emph{IEEE Robotics and Automation Letters}, 9\penalty0 (3):\penalty0 2543--2550, 2024.

\bibitem[Kahneman(2011)]{kahneman2011thinking}
Daniel Kahneman.
\newblock \emph{Thinking, fast and slow}.
\newblock macmillan, 2011.

\bibitem[Khazatsky et~al.(2024)Khazatsky, Pertsch, Nair, Balakrishna, Dasari, Karamcheti, Nasiriany, Srirama, Chen, Ellis, Fagan, Hejna, Itkina, Lepert, Ma, Miller, Wu, Belkhale, Dass, Ha, Jain, Lee, Lee, Memmel, Park, Radosavovic, Wang, Zhan, Black, Chi, Hatch, Lin, Lu, Mercat, Rehman, Sanketi, Sharma, Simpson, Vuong, Walke, Wulfe, Xiao, Yang, Yavary, Zhao, Agia, Baijal, Castro, Chen, Chen, Chung, Drake, Foster, Gao, Herrera, Heo, Hsu, Hu, Jackson, Le, Li, Lin, Lin, Ma, Maddukuri, Mirchandani, Morton, Nguyen, O'Neill, Scalise, Seale, Son, Tian, Tran, Wang, Wu, Xie, Yang, Yin, Zhang, Bastani, Berseth, Bohg, Goldberg, Gupta, Gupta, Jayaraman, Lim, Malik, Martín-Martín, Ramamoorthy, Sadigh, Song, Wu, Yip, Zhu, Kollar, Levine, and Finn]{khazatsky2024droid}
Alexander Khazatsky, Karl Pertsch, Suraj Nair, Ashwin Balakrishna, Sudeep Dasari, Siddharth Karamcheti, Soroush Nasiriany, Mohan~Kumar Srirama, Lawrence~Yunliang Chen, Kirsty Ellis, Peter~David Fagan, Joey Hejna, Masha Itkina, Marion Lepert, Yecheng~Jason Ma, Patrick~Tree Miller, Jimmy Wu, Suneel Belkhale, Shivin Dass, Huy Ha, Arhan Jain, Abraham Lee, Youngwoon Lee, Marius Memmel, Sungjae Park, Ilija Radosavovic, Kaiyuan Wang, Albert Zhan, Kevin Black, Cheng Chi, Kyle~Beltran Hatch, Shan Lin, Jingpei Lu, Jean Mercat, Abdul Rehman, Pannag~R Sanketi, Archit Sharma, Cody Simpson, Quan Vuong, Homer~Rich Walke, Blake Wulfe, Ted Xiao, Jonathan~Heewon Yang, Arefeh Yavary, Tony~Z. Zhao, Christopher Agia, Rohan Baijal, Mateo~Guaman Castro, Daphne Chen, Qiuyu Chen, Trinity Chung, Jaimyn Drake, Ethan~Paul Foster, Jensen Gao, David~Antonio Herrera, Minho Heo, Kyle Hsu, Jiaheng Hu, Donovon Jackson, Charlotte Le, Yunshuang Li, Kevin Lin, Roy Lin, Zehan Ma, Abhiram Maddukuri, Suvir Mirchandani, Daniel Morton, Tony Nguyen,
  Abigail O'Neill, Rosario Scalise, Derick Seale, Victor Son, Stephen Tian, Emi Tran, Andrew~E. Wang, Yilin Wu, Annie Xie, Jingyun Yang, Patrick Yin, Yunchu Zhang, Osbert Bastani, Glen Berseth, Jeannette Bohg, Ken Goldberg, Abhinav Gupta, Abhishek Gupta, Dinesh Jayaraman, Joseph~J Lim, Jitendra Malik, Roberto Martín-Martín, Subramanian Ramamoorthy, Dorsa Sadigh, Shuran Song, Jiajun Wu, Michael~C. Yip, Yuke Zhu, Thomas Kollar, Sergey Levine, and Chelsea Finn.
\newblock Droid: A large-scale in-the-wild robot manipulation dataset.
\newblock 2024.

\bibitem[Kim et~al.(2024)Kim, Pertsch, Karamcheti, Xiao, Balakrishna, Nair, Rafailov, Foster, Lam, Sanketi, et~al.]{kim2024openvla}
Moo~Jin Kim, Karl Pertsch, Siddharth Karamcheti, Ted Xiao, Ashwin Balakrishna, Suraj Nair, Rafael Rafailov, Ethan Foster, Grace Lam, Pannag Sanketi, et~al.
\newblock Openvla: An open-source vision-language-action model.
\newblock \emph{arXiv preprint arXiv:2406.09246}, 2024.

\bibitem[Li et~al.(2023{\natexlab{a}})Li, Li, Savarese, and Hoi]{li2023blip}
Junnan Li, Dongxu Li, Silvio Savarese, and Steven Hoi.
\newblock Blip-2: Bootstrapping language-image pre-training with frozen image encoders and large language models.
\newblock In \emph{International conference on machine learning}, pp.\  19730--19742. PMLR, 2023{\natexlab{a}}.

\bibitem[Li et~al.(2024)Li, Zhang, Geng, Geng, Long, Shen, Zhang, Liu, and Dong]{li2024manipllm}
Xiaoqi Li, Mingxu Zhang, Yiran Geng, Haoran Geng, Yuxing Long, Yan Shen, Renrui Zhang, Jiaming Liu, and Hao Dong.
\newblock Manipllm: Embodied multimodal large language model for object-centric robotic manipulation.
\newblock In \emph{Proceedings of the IEEE/CVF Conference on Computer Vision and Pattern Recognition}, pp.\  18061--18070, 2024.

\bibitem[Li et~al.(2023{\natexlab{b}})Li, Liu, Zhang, Yu, Xu, Wu, Cheang, Jing, Zhang, Liu, et~al.]{li2023vision}
Xinghang Li, Minghuan Liu, Hanbo Zhang, Cunjun Yu, Jie Xu, Hongtao Wu, Chilam Cheang, Ya~Jing, Weinan Zhang, Huaping Liu, et~al.
\newblock Vision-language foundation models as effective robot imitators.
\newblock \emph{arXiv preprint arXiv:2311.01378}, 2023{\natexlab{b}}.

\bibitem[Li et~al.(2025)Li, Chen, Liu, Wang, VS, Ji, Lan, Zhang, Zhao, Radhakrishnan, et~al.]{li2025eagle}
Zhiqi Li, Guo Chen, Shilong Liu, Shihao Wang, Vibashan VS, Yishen Ji, Shiyi Lan, Hao Zhang, Yilin Zhao, Subhashree Radhakrishnan, et~al.
\newblock Eagle 2: Building post-training data strategies from scratch for frontier vision-language models.
\newblock \emph{arXiv preprint arXiv:2501.14818}, 2025.

\bibitem[Lin et~al.(2018)Lin, Zhao, Yang, and Zhang]{lin2018episodic}
Zichuan Lin, Tianqi Zhao, Guangwen Yang, and Lintao Zhang.
\newblock Episodic memory deep q-networks.
\newblock \emph{arXiv preprint arXiv:1805.07603}, 2018.

\bibitem[Lipman et~al.(2022)Lipman, Chen, Ben-Hamu, Nickel, and Le]{lipman2022flow}
Yaron Lipman, Ricky~TQ Chen, Heli Ben-Hamu, Maximilian Nickel, and Matt Le.
\newblock Flow matching for generative modeling.
\newblock \emph{arXiv preprint arXiv:2210.02747}, 2022.

\bibitem[Liu et~al.(2024{\natexlab{a}})Liu, Zhu, Gao, Feng, Liu, Zhu, and Stone]{liu2024libero}
Bo~Liu, Yifeng Zhu, Chongkai Gao, Yihao Feng, Qiang Liu, Yuke Zhu, and Peter Stone.
\newblock Libero: Benchmarking knowledge transfer for lifelong robot learning.
\newblock \emph{Advances in Neural Information Processing Systems}, 36, 2024{\natexlab{a}}.

\bibitem[Liu et~al.(2024{\natexlab{b}})Liu, Yan, Zheng, Feng, Huang, and Ma]{liu2024robouniview}
Fanfan Liu, Feng Yan, Liming Zheng, Chengjian Feng, Yiyang Huang, and Lin Ma.
\newblock Robouniview: Visual-language model with unified view representation for robotic manipulation.
\newblock \emph{arXiv preprint arXiv:2406.18977}, 2024{\natexlab{b}}.

\bibitem[Liu et~al.(2024{\natexlab{c}})Liu, Li, Wu, and Lee]{liu2024visual}
Haotian Liu, Chunyuan Li, Qingyang Wu, and Yong~Jae Lee.
\newblock Visual instruction tuning.
\newblock \emph{Advances in neural information processing systems}, 36, 2024{\natexlab{c}}.

\bibitem[Liu et~al.(2024{\natexlab{d}})Liu, Liu, Wang, Lee, Zhou, An, Yang, Zhang, Guo, and Zhang]{liu2024robomamba}
Jiaming Liu, Mengzhen Liu, Zhenyu Wang, Lily Lee, Kaichen Zhou, Pengju An, Senqiao Yang, Renrui Zhang, Yandong Guo, and Shanghang Zhang.
\newblock Robomamba: Multimodal state space model for efficient robot reasoning and manipulation.
\newblock \emph{arXiv preprint arXiv:2406.04339}, 2024{\natexlab{d}}.

\bibitem[Lu et~al.(2024)Lu, Zhang, Wang, Liu, Lu, and Tang]{lu2024manigaussian}
Guanxing Lu, Shiyi Zhang, Ziwei Wang, Changliu Liu, Jiwen Lu, and Yansong Tang.
\newblock Manigaussian: Dynamic gaussian splatting for multi-task robotic manipulation.
\newblock In \emph{European Conference on Computer Vision}, pp.\  349--366. Springer, 2024.

\bibitem[Mees et~al.(2022)Mees, Hermann, Rosete-Beas, and Burgard]{mees2022calvin}
Oier Mees, Lukas Hermann, Erick Rosete-Beas, and Wolfram Burgard.
\newblock Calvin: A benchmark for language-conditioned policy learning for long-horizon robot manipulation tasks.
\newblock \emph{IEEE Robotics and Automation Letters (RA-L)}, 7\penalty0 (3):\penalty0 7327--7334, 2022.

\bibitem[Min et~al.(2023)Min, Pan, Dai, Kawsar, Li, and Wang]{min2023trajectory}
Chuan Min, Yongjun Pan, Wei Dai, Ibna Kawsar, Zhixiong Li, and Gengxiang Wang.
\newblock Trajectory optimization of an electric vehicle with minimum energy consumption using inverse dynamics model and servo constraints.
\newblock \emph{Mechanism and Machine Theory}, 181:\penalty0 105185, 2023.

\bibitem[O'Neill et~al.(2023)O'Neill, Rehman, Gupta, Maddukuri, Gupta, Padalkar, Lee, Pooley, Gupta, Mandlekar, et~al.]{o2023open}
Abby O'Neill, Abdul Rehman, Abhinav Gupta, Abhiram Maddukuri, Abhishek Gupta, Abhishek Padalkar, Abraham Lee, Acorn Pooley, Agrim Gupta, Ajay Mandlekar, et~al.
\newblock Open x-embodiment: Robotic learning datasets and rt-x models.
\newblock \emph{arXiv preprint arXiv:2310.08864}, 2023.

\bibitem[Pertsch et~al.(2025)Pertsch, Stachowicz, Ichter, Driess, Nair, Vuong, Mees, Finn, and Levine]{pertsch2025fast}
Karl Pertsch, Kyle Stachowicz, Brian Ichter, Danny Driess, Suraj Nair, Quan Vuong, Oier Mees, Chelsea Finn, and Sergey Levine.
\newblock Fast: Efficient action tokenization for vision-language-action models.
\newblock \emph{arXiv preprint arXiv:2501.09747}, 2025.

\bibitem[Pritzel et~al.(2017)Pritzel, Uria, Srinivasan, Badia, Vinyals, Hassabis, Wierstra, and Blundell]{pritzel2017neural}
Alexander Pritzel, Benigno Uria, Sriram Srinivasan, Adria~Puigdomenech Badia, Oriol Vinyals, Demis Hassabis, Daan Wierstra, and Charles Blundell.
\newblock Neural episodic control.
\newblock In \emph{International conference on machine learning}, pp.\  2827--2836. PMLR, 2017.

\bibitem[Radford et~al.(2021)Radford, Kim, Hallacy, Ramesh, Goh, Agarwal, Sastry, Askell, Mishkin, Clark, et~al.]{radford2021learning}
Alec Radford, Jong~Wook Kim, Chris Hallacy, Aditya Ramesh, Gabriel Goh, Sandhini Agarwal, Girish Sastry, Amanda Askell, Pamela Mishkin, Jack Clark, et~al.
\newblock Learning transferable visual models from natural language supervision.
\newblock In \emph{International conference on machine learning}, pp.\  8748--8763. PMLR, 2021.

\bibitem[Radosavovic et~al.(2023)Radosavovic, Xiao, James, Abbeel, Malik, and Darrell]{radosavovic2023real}
Ilija Radosavovic, Tete Xiao, Stephen James, Pieter Abbeel, Jitendra Malik, and Trevor Darrell.
\newblock Real-world robot learning with masked visual pre-training.
\newblock In \emph{Conference on Robot Learning}, pp.\  416--426. PMLR, 2023.

\bibitem[Reuss et~al.(2024)Reuss, Ömer Erdinç~Yağmurlu, Wenzel, and Lioutikov]{reuss2024multimodaldiffusiontransformerlearning}
Moritz Reuss, Ömer Erdinç~Yağmurlu, Fabian Wenzel, and Rudolf Lioutikov.
\newblock Multimodal diffusion transformer: Learning versatile behavior from multimodal goals, 2024.
\newblock URL \url{https://arxiv.org/abs/2407.05996}.

\bibitem[Shi et~al.(2025)Shi, Ichter, Equi, Ke, Pertsch, Vuong, Tanner, Walling, Wang, Fusai, et~al.]{shi2025hi}
Lucy~Xiaoyang Shi, Brian Ichter, Michael Equi, Liyiming Ke, Karl Pertsch, Quan Vuong, James Tanner, Anna Walling, Haohuan Wang, Niccolo Fusai, et~al.
\newblock Hi robot: Open-ended instruction following with hierarchical vision-language-action models.
\newblock \emph{arXiv preprint arXiv:2502.19417}, 2025.

\bibitem[Tian et~al.(2024)Tian, Yang, Zeng, Wang, Lin, Dong, and Pang]{tian2024predictive}
Yang Tian, Sizhe Yang, Jia Zeng, Ping Wang, Dahua Lin, Hao Dong, and Jiangmiao Pang.
\newblock Predictive inverse dynamics models are scalable learners for robotic manipulation.
\newblock \emph{arXiv preprint arXiv:2412.15109}, 2024.

\bibitem[Tulving(2002)]{tulving2002episodic}
Endel Tulving.
\newblock Episodic memory: From mind to brain.
\newblock \emph{Annual review of psychology}, 53\penalty0 (1):\penalty0 1--25, 2002.

\bibitem[Tulving et~al.(1972)]{tulving1972episodic}
Endel Tulving et~al.
\newblock Episodic and semantic memory.
\newblock \emph{Organization of memory}, 1\penalty0 (381-403):\penalty0 1, 1972.

\bibitem[Walke et~al.(2023)Walke, Black, Lee, Kim, Du, Zheng, Zhao, Hansen-Estruch, Vuong, He, Myers, Fang, Finn, and Levine]{walke2023bridgedata}
Homer Walke, Kevin Black, Abraham Lee, Moo~Jin Kim, Max Du, Chongyi Zheng, Tony Zhao, Philippe Hansen-Estruch, Quan Vuong, Andre He, Vivek Myers, Kuan Fang, Chelsea Finn, and Sergey Levine.
\newblock Bridgedata v2: A dataset for robot learning at scale, 2023.

\bibitem[Wen et~al.(2024)Wen, Lin, Zhu, Han, Xu, Zhao, and Liang]{wen2024vidman}
Youpeng Wen, Junfan Lin, Yi~Zhu, Jianhua Han, Hang Xu, Shen Zhao, and Xiaodan Liang.
\newblock Vidman: Exploiting implicit dynamics from video diffusion model for effective robot manipulation.
\newblock \emph{Advances in Neural Information Processing Systems}, 37:\penalty0 41051--41075, 2024.

\bibitem[Wu et~al.(2023)Wu, Jing, Cheang, Chen, Xu, Li, Liu, Li, and Kong]{wu2023unleashing}
Hongtao Wu, Ya~Jing, Chilam Cheang, Guangzeng Chen, Jiafeng Xu, Xinghang Li, Minghuan Liu, Hang Li, and Tao Kong.
\newblock Unleashing large-scale video generative pre-training for visual robot manipulation.
\newblock \emph{arXiv preprint arXiv:2312.13139}, 2023.

\bibitem[Xiang et~al.(2023)Xiang, Yang, Huang, and Wang]{xiang2023denoising}
Weilai Xiang, Hongyu Yang, Di~Huang, and Yunhong Wang.
\newblock Denoising diffusion autoencoders are unified self-supervised learners.
\newblock In \emph{Proceedings of the IEEE/CVF International Conference on Computer Vision}, pp.\  15802--15812, 2023.

\bibitem[Yang et~al.(2024)Yang, Teng, Zheng, Ding, Huang, Xu, Yang, Hong, Zhang, Feng, et~al.]{yang2024cogvideox}
Zhuoyi Yang, Jiayan Teng, Wendi Zheng, Ming Ding, Shiyu Huang, Jiazheng Xu, Yuanming Yang, Wenyi Hong, Xiaohan Zhang, Guanyu Feng, et~al.
\newblock Cogvideox: Text-to-video diffusion models with an expert transformer.
\newblock \emph{arXiv preprint arXiv:2408.06072}, 2024.

\bibitem[Ye et~al.(2024)Ye, Jang, Jeon, Joo, Yang, Peng, Mandlekar, Tan, Chao, Lin, et~al.]{ye2024latent}
Seonghyeon Ye, Joel Jang, Byeongguk Jeon, Sejune Joo, Jianwei Yang, Baolin Peng, Ajay Mandlekar, Reuben Tan, Yu-Wei Chao, Bill~Yuchen Lin, et~al.
\newblock Latent action pretraining from videos.
\newblock \emph{arXiv preprint arXiv:2410.11758}, 2024.

\bibitem[Yu et~al.(2020)Yu, Quillen, He, Julian, Hausman, Finn, and Levine]{yu2020meta}
Tianhe Yu, Deirdre Quillen, Zhanpeng He, Ryan Julian, Karol Hausman, Chelsea Finn, and Sergey Levine.
\newblock Meta-world: A benchmark and evaluation for multi-task and meta reinforcement learning.
\newblock In \emph{Conference on robot learning}, pp.\  1094--1100. PMLR, 2020.

\bibitem[Zhang et~al.(2023)Zhang, Han, Liu, Gao, Zhou, Hu, Yan, Lu, Li, and Qiao]{zhang2023llama}
Renrui Zhang, Jiaming Han, Chris Liu, Peng Gao, Aojun Zhou, Xiangfei Hu, Shilin Yan, Pan Lu, Hongsheng Li, and Yu~Qiao.
\newblock Llama-adapter: Efficient fine-tuning of language models with zero-init attention.
\newblock \emph{arXiv preprint arXiv:2303.16199}, 2023.

\bibitem[Zhang et~al.(2024{\natexlab{a}})Zhang, Jiang, Zhang, Lin, Guo, Qiu, Zhou, Lu, Chang, Gao, et~al.]{zhang2024mathverse}
Renrui Zhang, Dongzhi Jiang, Yichi Zhang, Haokun Lin, Ziyu Guo, Pengshuo Qiu, Aojun Zhou, Pan Lu, Kai-Wei Chang, Peng Gao, et~al.
\newblock Mathverse: Does your multi-modal llm truly see the diagrams in visual math problems?
\newblock \emph{ECCV 2024}, 2024{\natexlab{a}}.

\bibitem[Zhang et~al.(2024{\natexlab{b}})Zhang, Wei, Jiang, Guo, Li, Zhang, Tong, Liu, Zhou, Wei, et~al.]{zhang2024mavis}
Renrui Zhang, Xinyu Wei, Dongzhi Jiang, Ziyu Guo, Shicheng Li, Yichi Zhang, Chengzhuo Tong, Jiaming Liu, Aojun Zhou, Bin Wei, et~al.
\newblock Mavis: Mathematical visual instruction tuning with an automatic data engine.
\newblock \emph{arXiv preprint arXiv:2407.08739}, 2024{\natexlab{b}}.

\end{thebibliography}
